\documentclass{article}

\usepackage[main, final]{neurips_2026}

\usepackage[utf8]{inputenc} 
\usepackage[T1]{fontenc}    
\usepackage{hyperref}       
\usepackage{url}            
\usepackage{booktabs}       
\usepackage{amsfonts}       
\usepackage{nicefrac}       
\usepackage{microtype}      
\usepackage[table]{xcolor}         
\usepackage{amsmath}
\usepackage{amssymb}
\usepackage{mathtools}
\usepackage{amsthm}
\usepackage{multirow}
\usepackage{graphicx}
\usepackage{subcaption}
\usepackage{pifont}
\usepackage{hyperref}
\usepackage[capitalize,noabbrev]{cleveref}
\usepackage{wrapfig}
\usepackage{enumitem}

\title{FIND: Frequency Invariance Disentanglement for Test-Time Adaptation in LiDAR 3D Detection}

\author{%
  Dongxiao Li \\
  School of Computer Science\\
  Peking University\\
  Beijing 100871, China\\
  \texttt{dxli25@stu.pku.edu.cn}
  \And
  Zixuan Hu \\
  School of Computer Science\\
  Peking University\\
  Beijing 100871, China\\
  \texttt{hzxuan@pku.edu.cn}
  \And
  Ling-Yu Duan\thanks{Corresponding author.} \\
  School of Computer Science\\
  Peking University\\
  Beijing 100871, China\\
  \texttt{lingyu@pku.edu.cn}
}

\begin{document}

\setcounter{footnote}{1}
\maketitle
\setcounter{footnote}{0}

\begin{abstract}
LiDAR-based 3D Object Detection (L3OD) is a fundamental 3D perception task, yet its performance is often compromised by domain shifts arising from diverse sensor configurations and environmental conditions.
Existing Test-Time Adaptation (TTA) methods primarily leverage self-supervision in the spatial domain, yet they frequently confound semantic content with domain style due to the sparse, non-Euclidean nature of point clouds.
To overcome this limitation, we propose \textbf{FIND} (\textbf{F}requency \textbf{IN}variance \textbf{D}isentanglement), a novel TTA framework that precisely disentangles domain-invariant content from domain-specific interference within the frequency domain.
Our methodology utilizes B-Spline fitting to construct locally adaptive filters integrated into a dual-stream architecture: an Invariant Stream extracts robust features to guide pseudo-labeling, while a Specific Stream enforces consistency against domain perturbations.
By combining invariance-driven alignment with stability-guided regularization, our approach dynamically extracts robust domain-invariant features while suppressing domain-specific interference.
Extensive experiments on cross-dataset adaptation and robust corruption benchmarks demonstrate that FIND significantly outperforms state-of-the-art methods.
The source code is publicly available at \url{https://github.com/llldx03/FIND}.
\end{abstract}
\section{Introduction}

\begin{figure}[!t]
    \centering
    \includegraphics[width=\linewidth]{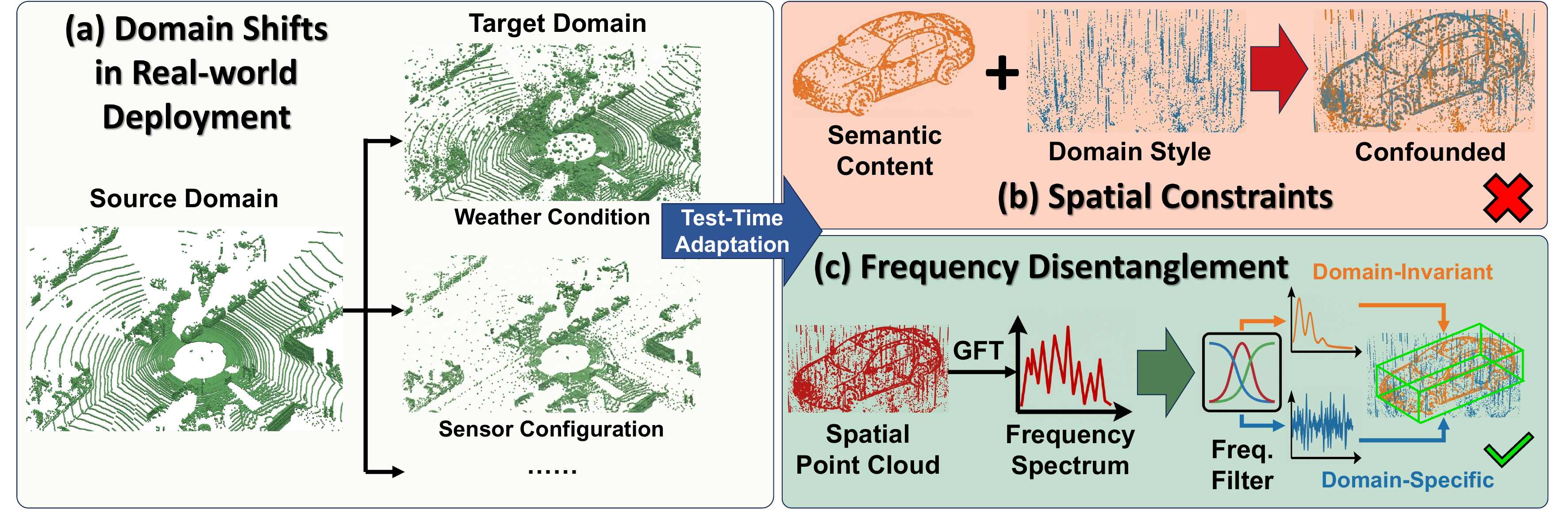}
    \caption{(a) Illustration of domain shift types in L3OD. (b) Spatial constraints of point clouds, i.e., complex entanglement. (c) Disentanglement paradigm through Fourier transformation.}
    \vspace{-0.5cm}
    \label{fig:teaser}
\end{figure}

LiDAR-based 3D Object Detection (L3OD)~\cite{pointpillars, li2022uvtr, mssvt++} is a fundamental perception task aiming to estimate the spatial locations and semantic categories of objects from sparse 3D point clouds.
While prior L3OD studies have achieved significant progress, they typically rely on the assumption that training (source) and test (target) data are identically distributed~\cite{second, pvrcnn, wang2023dsvt}.
However, as shown in Fig.~\ref{fig:teaser}(a), real-world deployment often involves variations such as different sensor configurations or diverse weather conditions, causing domain shifts~\cite{3drobust}.
These shifts render the target domain out-of-distribution (OOD), significantly degrading detector performance~\cite{st3d++, hu2023density, shin2025diffrefine}.

To tackle the domain shift issue efficiently, Test-Time Adaptation (TTA) has emerged as a critical paradigm, enabling model adaptation from source to target domains during inference~\cite{tent, dda, tta1}.
Conventional TTA research has achieved significant success in 2D vision by leveraging self-supervision such as entropy minimization~\cite{tent, eata} and pseudo labeling~\cite{cotta, Dbler2022RobustMT}.
Several recent works extend those techniques from 2D tasks to L3OD, aiming to bridge the domain gap within the spatial domain (i.e., 3D coordinate system) by enforcing consistency via spatial perturbations~\cite{reg-tta} or model ensembles~\cite{mos}.
However, given the sparse and non-Euclidean nature of LiDAR data~\cite{Ahmed2018EdgeAC}, current spatial methods often confound domain-specific styles with essential semantic content, as shown in Fig.~\ref{fig:teaser}(b). Inspired by Fourier-based analysis~\cite{fda}, we delve into the frequency domain to break these spatial constraints, capturing the geometric topology from a compact spectral representation as shown in Fig.~\ref{fig:teaser}(c).

Contrary to common assumptions that domain-specific components are solely isolated in high-frequency bands~\cite{highpass, he2024towards}, our empirical analysis reveals a pervasive distribution across the full spectrum.
As shown in Fig.~\ref{fig:obser}(a), the rigid band-level filtering inevitably discards critical geometric structures, which severely compromises semantic consistency. 
Furthermore, Fig.~\ref{fig:obser}(b) highlights that domain discrepancies exhibit significant fluctuations across frequencies.
These rapid variations imply that the entanglement of domain-specific components exhibits abrupt changes between adjacent frequency bands, rendering global or smooth filters ineffective.
To resolve this complex entanglement,
it is essential to devise a filtering mechanism capable of independent local modulation, allowing for precise disentanglement without inducing global spectral distortion.

Motivated by these observations, we propose \textbf{F}requency \textbf{IN}variance \textbf{D}isentanglement (\textbf{FIND}), the first TTA framework designed to disentangle 3D point cloud representations in the frequency domain.
Specifically, we exploit the local support property of B-Spline fitting~\cite{bspline} to construct filtering modules, that afford fine-grained, band-selective control. Integrated into a dual-stream architecture, this design establishes an invariance-driven alignment:an Invariant Stream extracts robust spectrally stable components to generate reliable pseudo-labels, while a Specific Stream enforces consistency against domain-specific perturbations.
To further align the disentanglement with the distribution of invariant components, we incorporate a stability-guided module, which quantifies the Coefficient of Variation (CV) of the spectral energy, serving as a data-driven indicator of per-band invariance.
Through this synergetic integration of spectral filtering and dual-stream learning, FIND effectively disentangles spectrally stable representations from domain-sensitive variations, facilitating adaptation to unseen domains.

We evaluate the effectiveness and efficiency of our method through experiments on cross-dataset adaptation tasks (nuScenes/Waymo$\rightarrow$KITTI), achieving SOTA results with average improvements of \textbf{+2.27} $\text{AP}_{\text{BEV}}$ and \textbf{+3.24} $\text{AP}_{\text{3D}}$.
Furthermore, we also evaluate its performance on a wide range of real-world corruptions (KITTI-C), improving \textbf{+2.14} $\text{AP}_{\text{3D}}$ averagely compared to existing approaches.

\textbf{Contributions:}
(1) To the best of our knowledge, we pioneer frequency disentanglement in L3OD by proposing the first TTA framework that isolates domain-invariant components, mitigating the performance degradation issue under domain shifts.
(2) We propose a dual-stream disentangled learning architecture incorporating B-Spline frequency filters, leveraging their local support properties to capture narrow-band spectral components.
(3) We propose a disentangled learning strategy that integrates dual-stream alignment and stability-guided regularization, establishing a virtuous cycle between the L3OD model and the disentanglement filters.
\section{Related Work}
\label{related}
\subsection{Test-Time Adaptation}
Test-Time Adaptation (TTA) aims to enhance model performance on OOD samples during inference. Existing TTA methods are categorized into two groups based on their interaction with the source training process: Test-Time Training \cite{ttt, ttt++, liu2024depth} and Fully Test-Time Adaptation \cite{memo, deyo, recap}. In this paper, we focus on Fully TTA, which adapts the model exclusively during the testing phase without altering the source training objective.
Following TENT~\cite{tent}'s use of entropy minimization for batch normalization adaptation, several self-supervised methods address domain shifts: CoTTA~\cite{cotta} employs mean-teacher to generate reliable pseudo-labels, while SAR~\cite{sar} minimizes sharpness-aware entropy for stability.
With the rising prominence of 3D vision, recent works have extended TTA paradigms to various 3D tasks including segmentation~\cite{d3ctta, pcseg}, navigation~\cite{kim2025testtime} and detection~\cite{monotta, dpo, duo}.
For L3OD, Reg-TTA3D~\cite{reg-tta} improves bounding box regression via confidence-guided refinement, while MOS~\cite{mos} aggregates historical checkpoints to leverage collective knowledge.
While current 3D TTA approaches struggle with entangled geometric structure and sensor noise in the spatial domain, our method exploits the frequency domain to disentangle domain-invariant components.

\subsection{Graph Spectral Analysis}
Graph Spectral Analysis (GSA) extends classical signal processing theories to non-Euclidean domains~\cite{gft}, offering a robust mathematical framework for analyzing irregular data structures such as 3D point clouds~\cite{gsp}. By performing eigen-decomposition on the Graph Laplacian, GSA maps spatial signals to the spectral domain to uncover underlying geometric topology.
In the realm of 3D vision, GSA has been widely applied to tasks including point cloud denoising~\cite{pcdenoise}, compression~\cite{frpc, pccompress}, and geometric feature learning~\cite{pcfeature}. Leveraging the compactness of spectral representations~\cite{spectralgan}, frequency filters are employed to smooth geometric surfaces or suppress high-frequency noise~\cite{pcattack, liu2023point}.
Despite advances, analysis of spectral component variations under domain shift is still inadequacy and inaccuracy.
\section{Preliminary}
\label{sec:3}
\textbf{Graph Fourier Transform (GFT)} enables the mapping of non-Euclidean point cloud data into the frequency domain. Given point clouds $X\in\mathbb{R}^{n\times3}$ consisting of $n$ points $\{x^{(i)}\}_{i=1}^n$, a graph is constructed via $k$-nearest neighbors (kNN) method, where edge weights are defined by a Gaussian kernel function:
\begin{equation}
    A_{i,j} = \exp\left(-\frac{\|x^{(i)} - x^{(j)}\|_2^2}{2\sigma^2}\right),
\end{equation}
where $\sigma$ is a kernel width parameter. The symmetric normalized Laplacian matrix serves as the fundamental operator in the frequency domain:
\begin{equation}
    L_{\text{sym}}=I-D^{-\frac{1}{2}}AD^{-\frac{1}{2}},
\end{equation}
where $D$ and $A$ denote the degree matrix and graph adjacency matrix respectively. The eigenvalues of $L_{\text{sym}}$, ordered as $0=\lambda_1\le\lambda_2\le\cdots\le\lambda_n\le 2$, correspond to spectral frequencies, where smaller and larger eigenvalues represent low and high frequencies, respectively. The eigenvector matrix $U$ forms the Graph Fourier basis, allowing the transformation of $X$ into the frequency domain via $\hat X=U^T X$. For each eigenvalue $\lambda_k$, the spectral energy is defined by
\begin{equation}
    \mathcal{E}_{\lambda_k}(X)=\|u_k^TX\|_2^2, 
\end{equation}
where $u_k$ is the corresponding eigenvector of $\lambda_k$.
The filter $H(\Lambda)$ can be directly applied in the frequency domain as $\hat X_{\text{filt}}= H(\Lambda) \hat X$, where the filter kernel $H(\Lambda) = \text{diag}[h(\lambda_k)]$ is a function of the diagonal eigenvalue matrix $\Lambda$. Finally, the signal is reconstructed in the vertex domain via Inverse GFT, formulated as $X_{\text{filt}} = U \hat{X}_{\text{filt}}$.

\textbf{B-spline fitting} utilizes piecewise polynomials with compactly supported basis functions to ensure local control. Unlike global interpolation, this structure confines parameter influence to specific regions, enhancing numerical stability and flexibility.
In the context of frequency filtering, this property allows for the independent optimization of local frequency bands.
Formally, given a specified interval $[a,b]$ and $(N+1)$ control points $(x_0,y_0), \dots, (x_N,y_N)$ where $a=x_0 < \dots < x_N=b$, a $K$-th degree spline function based on B-spline bases can be constructed as $S(x) = \sum_{i=0}^{N+K-1} c_i B_{i,K}(x),$
where the basis functions $B_{i,K}(x)$ (the $i$-th B-spline of degree $K$) are defined by the Cox-de Boor recursion formula~\cite{bspline}, detailed in Appendix ~\ref{appendix:b-spline}.
The control coefficients $c_i$ can be determined linearly through control point conditions and boundary derivative constraints.

\textbf{TTA setting} aims to address the challenge of distribution shifts by enabling a pre-trained model to adapt to the target distribution during inference, without the need for labeled data. Unlike traditional training on fixed datasets, TTA operates in an online manner, where the model $h_{\theta}$ with parameter set $\theta$ produces detection outputs while concurrently updating its parameters based on incoming test data.
Specifically, for L3OD task, given an input point cloud $X_t \in \mathbb{R}^{n \times 3}$, the goal is to generate accurate 3D bounding boxes $\{\mathcal{B}_t^{(i)}\}_{i=1}^{m}$ and semantic labels $\{\mathcal{C}_t^{(i)}\}_{i=1}^{m}$ for objects present in the scene, where $m$ denotes the number of objects. Each bounding box is typically parameterized as $\mathcal{B}^{(i)} = (\mathcal{P}^{(i)},\, \mathcal{D}^{(i)},\, \mathcal{O}^{(i)})$,
where $\mathcal{P}^{(i)} \in \mathbb{R}^3$ denotes the 3D center position, $\mathcal{D}^{(i)} \in \mathbb{R}^3$ represents the shape dimensions, and $\mathcal{O}^{(i)} \in [-\pi, \pi]$ encodes the orientation. Consequently, TTA process can be described by the following formula:
\begin{equation}
    \{\mathcal{B}_t\}, \{\mathcal{C}_t\}\leftarrow h_{\theta}(X_t),\; \theta\leftarrow\theta-\nabla\mathcal{L}_{\text{TTA}}(X_t),
\label{eq:tta}
\end{equation}
where $\mathcal{L}_{\text{TTA}}$ denotes the loss function used for self-training during adaptation. 

\begin{figure*}[!t]
    \centering
    \includegraphics[width=\linewidth]{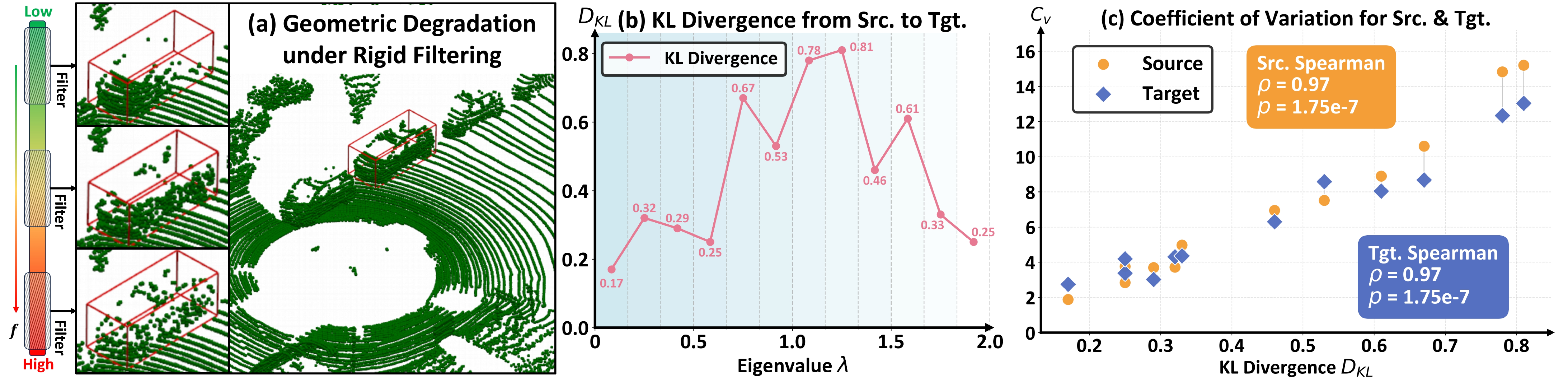}
    \caption{Spectral entanglement analysis of 3D point clouds. \textit{(a)} shows compromised geometric integrity resulting from rigid filtering distinct frequency bands. \textit{(b)} records cross-domain KL divergence ($D_{KL}$) across 12 uniform frequency bins. \textit{(c)} investigates Spearman correlation between Coefficient of Variation ($C_{v}$) and KL divergence within each bin for both source and target domains. Experiments are conducted on nuScenes $\rightarrow$ KITTI scenario. More empirical analysis on other scenarios are provided in Appendix~\ref{appendix:empirical}.}
    \label{fig:obser}
\end{figure*}

\section{Observation}
\label{sec:4}
To investigate 3D point cloud entanglement under domain shift, we conduct a spectral analysis comparing source (i.e., nuScenes) and target (i.e., KITTI) domains.
Leveraging GFT decomposition, we observe the global pervasiveness and local volatility of entangling properties in the frequency domain, showing inadequacy and inaccuracy of general filters.
It directly motivates our design of a geometry-reliable, local-sensitive and trend-aware filtering mechanism.

\textbf{Observation 1:} \textit{Rigid filtering compromises geometric integrity.}
Fig.~\ref{fig:obser}(a) demonstrates that ideal band-stop filtering damages key semantic content regardless of the selected range: removing low frequencies destroys global shape, mid-frequency suppression causes fragmentation, and high-frequency elimination erases fine details (e.g., wheels).
Controlled empirical analysis with fixed geometry further support this cross-band entanglement (Appendix~\ref{appendix:empirical}).

\textbf{Observation 2:} \textit{Filtering with global control is insensitive to frequency-dependent fluctuations.}
We quantify the per-band domain gap via KL divergence of spectral energy distributions:
\begin{equation}
    D_{KL}^{(i)}=\int_{0}^{+\infty} P_S(\mathcal{E}_i)\log\frac{P_S(\mathcal{E}_i)}{P_T(\mathcal{E}_i)}d\mathcal{E}_i,
\end{equation}
where $\mathcal{E}_i$ denotes spectral energy at band $[\lambda_{i-1},\lambda_i)$.
As shown in Fig.~\ref{fig:obser}(b), $D_{KL}$ persists across the entire spectrum but fluctuates sharply between neighboring bands.
Such narrow-band variations cannot be captured by globally controlled filters, necessitating fine-grained, band-selective spectral adjustment.

\textbf{Observation 3:} \textit{Intra-domain spectral instability correlates with cross-domain divergence.}
While directly minimizing $D_{KL}$ would be ideal, it is infeasible in TTA where source data is unavailable.
We examine the Coefficient of Variation ($C_v{=}\sigma/\mu$) of spectral energy within each band as an intra-domain alternative.
Fig.~\ref{fig:obser}(c) reveals a positive Spearman correlation between $C_v$ and $D_{KL}$, which can be attributed to frequency-specific susceptibility to environmental variation~\cite{miao2024improving,feng2025generalized}: bands sensitive to micro-perturbations within a domain (high $C_v$) tend to be more vulnerable to macroscopic sensor differences across domains (high $D_{KL}$).
This suggests that $C_v$ can serve as a source-free trend indicator for guiding spectral disentanglement during online adaptation.

\section{Methodology}

Based on the above observations, we propose a TTA method \textbf{F}requency \textbf{IN}variance \textbf{D}isentanglement (\textbf{FIND}) framework for L3OD.
We employ a dual-stream architecture with learnable B-Spline filters to disentangle \textit{domain-invariant} frequency components (semantic content) from \textit{domain-specific} ones (domain style, sampling patterns).

\subsection{Dual-Stream Architecture}
\label{sec:5.1}
To achieve this disentanglement, we construct a dual-stream architecture consisting of two identical models with distinct input filtering modules, called \textit{Invariant Stream} and \textit{Specific Stream}.
Filtering module processes point clouds via GFT and frequency filters.

\textbf{Semantic-Aware Graph Construction.}
To prioritize foreground objects over background noise, we construct graphs using point clouds constrained within predicted bounding boxes.
Specifically, for a point set $X$ within a bounding box, we center the coordinates (i.e., $X\leftarrow X-\bar{X}$) and construct a kNN graph following the formulation in Section~\ref{sec:3} to capture multi-scale geometric topology.
We then calculate the Laplacian spectrum, yielding a frequency-domain representation with semantic reliability, where multi-scale geometric features are mapped into distinct, localized narrow frequency bands. Capturing such features requires a filtering mechanism capable of local, fine-grained control.

\textbf{Spectral B-Spline Filter Design.}
To achieve precise frequency disentanglement, we exploit the local support property of B-Splines, where each basis function is active only within a compact interval, enabling a flexible, band-selective filter design.
Based on Sec.~\ref{sec:3}, we model the spectral response function $h(\lambda)$ as a linear combination of B-Spline basis functions. By dividing the spectral range $[0, 2]$ into $N$ bins, the filter is defined as:
\begin{equation}
    \label{equ:filter}
    h(\lambda) = [B_{0,K}(\lambda), B_{1,K}(\lambda), \cdots,B_{N+K-1,K}(\lambda)]\mathbf{C},
\end{equation}
where $B_{i,K}$ is B-Spline basis function of degree $K$, while $\mathbf{C}=[c_0,c_1,\cdots,c_{N+K-1}]^T \in \mathbb{R}^{N+K}$ are control parameters.
To determine the control vector $\mathbf{C} $, we impose a unified set of linear constraints that simultaneously enforce learnable response $\mathbf{\Delta}=[\delta_0,\delta_1,\cdots,\delta_N]$ at frequency bin endpoints and vanishing derivatives at the boundary:
\begin{equation}
    \mathbf{C} = [\mathbf{B}, \mathbf{D}]^{-T} [\mathbf{\Delta},\mathbf{0}_{K-1}]^T,
    \label{equ:sol}
\end{equation}
where $\mathbf{B} \in \mathbb{R}^{(N+K)\times(N+1)}$ with entries $\mathbf{B}_{ij} = B_{i,K}(\lambda_j)$ implements the interpolation constraints $h(\lambda_j)=\delta_j$, and $\mathbf{D} \in \mathbb{R}^{(N+K)\times(K-1)}$ with $\mathbf{D}_{ij} = B^{(j+1)}_{i,K}(\lambda_0)$ enforces the first $(K-1)$-th-order derivatives to be zero.
During online adaptation, we optimize the endpoint responses $\mathbf{\Delta}$ and obtain $\mathbf{C}$ through the differentiable linear mapping in Eq.~(\ref{equ:sol}). The knot vector, basis functions, boundary constraints, $N$, and $K$ remain fixed.
This enables the filter to learn arbitrary band-pass or band-stop responses, allowing for the precise extraction of target components from narrow frequency bands.

\textbf{Filter Initialization.}
Before deployment, we initialize the Invariant Stream using spectral statistics from the source domain. We partition the frequency spectrum into $N$ eigenvalue bins and compute the energy mean $\mu^{(s)}_i$ and standard deviation $\sigma^{(s)}_i$ for each bin $i$.
Based on \textbf{Obs. 3} in Sec.~\ref{sec:4}, since the Coefficient of Variation $C_v=\sigma/\mu$ of the energy distribution within each bin reflects the trend of domain-invariant property, we define the filter response $\delta_{i,\text{in}}\in[0,1]$ via an exponential decay function negatively correlated with instability:
\begin{equation}
    \label{equ:init}
    \delta_{i,\text{in}}=
    \begin{cases}
        \exp{\left(-\dfrac{\sigma_i^{(s)}/\mu_i^{(s)}}{\tau}\right)}, & i=1,\cdots, N,\\
        1, & i=0,
    \end{cases}
\end{equation}
where $\tau$ is a decay scale factor.
The continuous filter $h_{\text{in}}(\lambda)$ with control parameter $\mathbf{C_{\text{in}}}$ is subsequently derived via Eq.~(\ref{equ:filter}) and~(\ref{equ:sol}). Conversely, to capture domain-specific variations, we initialize the Specific Stream filter $H_{\text{sp}}(\Lambda)$ as a smooth high-pass filter controlled by $\delta_{i,\text{sp}}=(\lambda_i/2)^{K}$, preventing interference from the source domain.
Such a mild inductive bias facilitates effective learning of domain-specific disentanglement for the target domain in the subsequent stage.
The initialized filter thus encodes a source spectral prior, while subsequent online refinement uses only the incoming target stream, without accessing source samples or labels.

\begin{figure*}[!t]
    \centering
    \includegraphics[width=\linewidth]{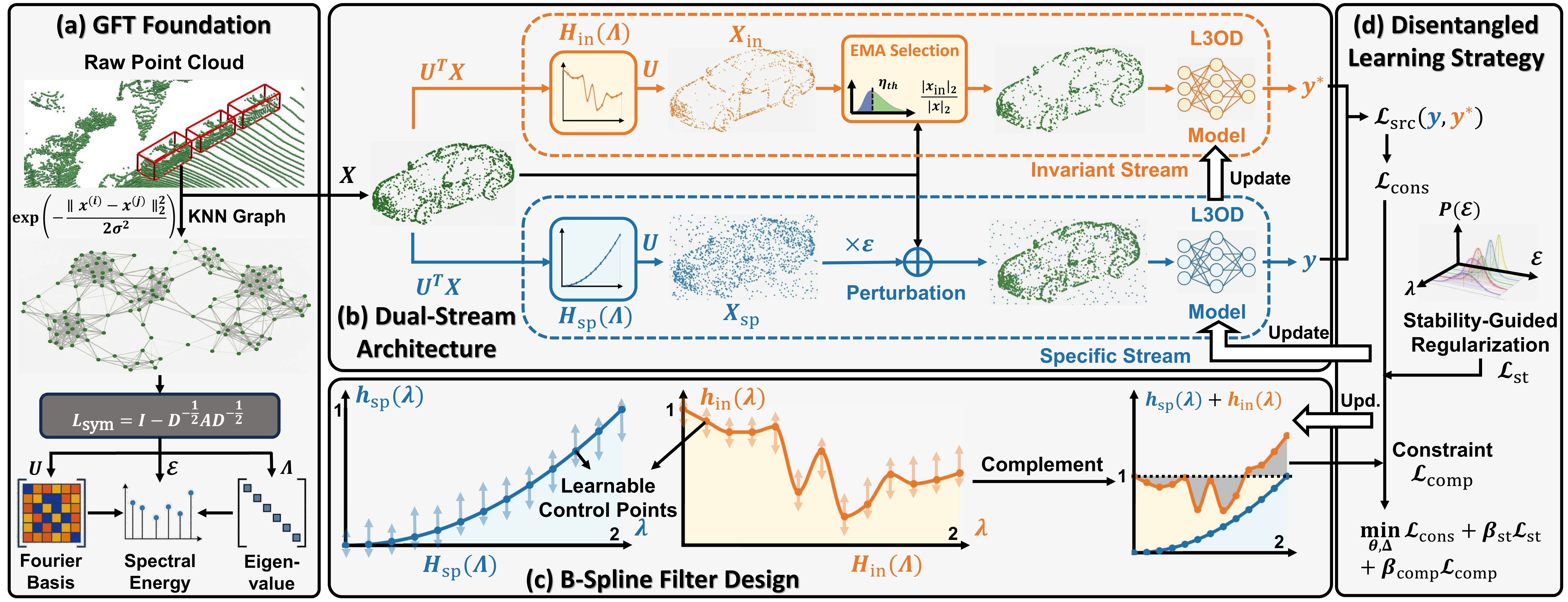}
    \caption{Overview of our FIND method. (a) GFT foundation for spectral analysis of point cloud. (b) Dual-stream architecture with distinct filter modules to disentangle domain-invariant semantic content (Sec.~\ref{sec:5.1}). (c) FIND employs B-Spline frequency filters to capture narrow-band invariant and specific components. (d) Through the disentangled learning strategy (Sec.~\ref{sec:5.2}), L3OD model and filtering modules can be optimized jointly to foster a virtuous cycle.}
    \label{fig:pipeline}
\end{figure*}

\subsection{Disentangled Learning Strategy}
\label{sec:5.2}
With the complementary dual-stream architecture established, we propose a disentangled learning strategy to jointly optimize the model parameters and filter coefficients during the adaptation phase.

\textbf{Invariance-Driven Alignment.}
In our proposed dual-stream framework, inputs to the Invariant Stream are filtered to yield domain-invariant representations.
Specifically, given a centered point cloud $X$, domain-invariant components are extracted via $X_{\text{in}}=UH_{\text{in}}(\Lambda)U^TX$.
To remove most domain-specific components and provide a reliable representation, we apply the following binary mask for point selection:
\begin{equation}
    \mathcal{M}_{\text{in}}(i)=\mathbb{I}\left(x^{(i)}\in X\left|\frac{\|x^{(i)}_\text{in}\|_2}{\|x^{(i)}\|_2} >\eta_{th}^{(t)}\right.\right),
\end{equation}
where $x^{(i)}\in\mathbb{R}^3$ denotes the $i$-th row of $X$ (i.e., the coordinate of the $i$-th point) and $\eta_{th}^{(t)}$ is a dynamic threshold updated via Exponential Moving Average (EMA) by
\begin{equation}
    \eta_{th}^{(t+1)}=\beta\dfrac{\sum_{i=1}^n \mathcal{M}_{\text{in}}(i)\|x^{(i)}_\text{in}\|_2 / \|x^{(i)}\|_2}{\sum_{i=1}^n \mathcal{M}_{\text{in}}(i)}+(1-\beta)\eta_{th}^{(t)},
\end{equation}
where $\beta$ is a moving average factor set to 0.001 in default, and $\eta_{th}^{(0)}$ is initialized as 0.2.
Applying $\mathcal{M}_{\text{in}}$, the input to the Invariant Stream is formulated as
\begin{equation}
\begin{aligned}
    \left[
    \begin{array}{cccc}
        x^{(i_1)}, &
        x^{(i_2)}, &
        \cdots, &
        x^{(i_M)}
    \end{array}
    \right]^T,\;
    \{i_1,i_2,\cdots,i_M\}=\{i|\mathcal{M}_{\text{in}}(i)=1\}.
\end{aligned}
\end{equation}
Simultaneously, the Specific Stream processes inputs by subsequently augmenting them with filtered domain-specific perturbations.
We extract the domain-specific components as $X_{\text{sp}}=UH_{\text{sp}}(\Lambda)U^TX$, and define the augmented input as $X+\epsilon_r X_{\text{sp}}$, where $\epsilon_r$ is a random coefficient sampled from $(-1,1)$.
The output of Invariant Stream serves as pseudo-labels for Specific Stream, formulating the consistency loss as
\begin{equation}
    \mathcal{L}_{\text{cons}}=\mathcal{L}_{\text{src}}(y, y^*),
\end{equation}
where $\mathcal{L}_{\text{src}}$ denotes the original loss function of the source model, while $y$ and $y^*$ represent the outputs of Specific Stream and Invariant Stream, respectively. 
By enforcing consistency between these parallel streams, the model effectively learns to reduce sensitivity to domain-specific interference.

\textbf{Online Stability-Guided Regularization.}
To accurately approximate domain-specific distribution of the target domain, we leverage online statistics to guide filter regularization.
To guide filter regularization, we accumulate spectral statistics solely from the incoming target batches. These statistics define the stability mask $S(\lambda)$, refining the filters initialized before deployment without requiring source data during adaptation.
Specifically, we record total sample size $N_{\text{tot}}$, first and second-order raw moment $\mathbb{E}(\mathcal{E}_i),\mathbb{E}(\mathcal{E}_i^2)$ of spectral energy $\mathcal{E}_i$ within the $i$-th bin of frequency domain for each test batch.
Upon receiving a new batch, we update the current statistics as follows:
\begin{equation}
\begin{aligned}
    \mathbb{E}(\mathcal{E}_i) \leftarrow \frac{N_{\text{tot}}\cdot\mathbb{E}(\mathcal{E}_i) + N^{(t)}\cdot \mathbb{E}^{(t)}(\mathcal{E}_i)}{N_{\text{tot}}+N^{(t)}},\;\;\;
    \mathbb{E}(\mathcal{E}_i^2) \leftarrow \frac{N_{\text{tot}}\cdot\mathbb{E}(\mathcal{E}_i^2) + N^{(t)}\cdot \mathbb{E}^{(t)}(\mathcal{E}_i^2)}{N_{\text{tot}}+N^{(t)}},\\
\end{aligned}
\end{equation}
where the superscript $^{(t)}$ denotes the statistics for the current batch.
Therefore, the mean $\tilde\mu_i^{(t)}=\mathbb{E}(\mathcal{E}_i)$ and standard deviation $\tilde{\sigma}_i^{(t)}=\sqrt{\mathbb{E}(\mathcal{E}_i^2)-\mathbb{E}^2(\mathcal{E}_i)}$ can be calculated.
Subsequently, the stability mask $S(\lambda)$ is constructed as
\begin{equation}
    \label{equ:stability}
    S(\lambda)=\sum_{i=1}^N \exp{\left(-\dfrac{\tilde\sigma_i^{(t)}/\tilde\mu_i^{(t)}}{\tau}\right)}\cdot\mathbb{I}(\lambda\in[\lambda_{i-1}, \lambda_{i})),
\end{equation}
whose form is consistent with Equ.~(\ref{equ:init}).
To guide domain-specific disentanglement, we update filter parameters $\mathbf{\Delta}_{\text{in}}$, $\mathbf{\Delta}_{\text{sp}}$ by the regularization term $\mathcal{L}_{\text{st}}$ with the stability mask:
\begin{equation}
    \mathcal{L}_{\text{st}} = \int_0^2 \left[h_{\text{in}}(\lambda) \cdot (1 - S(\lambda)) + h_{\text{sp}}(\lambda) \cdot S(\lambda)\right] \, d\lambda,
\end{equation}
where $S(\lambda)$ serves as an adaptive weighting factor for filter optimization.

\textbf{Complementary Constraint.}
To preclude trivial solutions and enforce the opposition between domain-invariant and domain-specific components, we impose a complementary constraint requiring the dual-stream filters to sum to unity:
\begin{equation}
    \mathcal{L}_{\text{comp}} = \int_0^2 |h_{\text{in}}(\lambda)+h_{\text{sp}}(\lambda)-1|^2\, d\lambda.
\end{equation}

\subsection{Overall Objective}
The overall objective unifies consistency, stability and complementarity into a single framework, jointly optimizing the model parameters $\theta$ and filter endpoint responses $\mathbf{\Delta}=\{\mathbf{\Delta}_{\text{in}},\mathbf{\Delta}_{\text{sp}}\}$ within a dual-stream architecture.
Gradients propagate through Eq.~(\ref{equ:sol}) to $\mathbf{\Delta}$, while $\mathbf{C}_{\text{in}}$ and $\mathbf{C}_{\text{sp}}$ are derived rather than independently optimized.
This joint optimization fosters a virtuous cycle: filters disentangle domain-invariant features to provide reliable guidance, while enhanced model robustness concurrently refines filter parameters to distinguish semantic content from domain-specific interference.
The overall objective is as follows:
\begin{equation}
    \min_{\theta,\mathbf{\Delta}} \mathcal{L}_{\text{cons}} +\beta_{\text{st}}\mathcal{L}_{\text{st}} +\beta_{\text{comp}}\mathcal{L}_{\text{comp}},
\end{equation}
where $\beta_{\text{st}}$ and $\beta_{\text{comp}}$ are balance factors set to $0.8$ and $0.6$ by default respectively.

\begin{table*}[t]
    \centering
    \caption{Comparisons with state-of-the-art methods on nuScenes$\rightarrow$KITTI and Waymo$\rightarrow$KITTI scenarios. We report {$\text{AP}_{\text{BEV}} / \text{AP}_{\text{3D}}$} of Car category in moderate difficulty. "Source" and "Oracle" denote model performance with no adaptation and fully supervised training, respectively. We highlight the best and second results with \textbf{bold} and \underline{underline} respectively. Due to the space limit, the complete results with more scenarios and categories are provided in Tab.~\ref{tab:all-cross} of Appendix.~\ref{appendix:exp}.}
    \resizebox{\textwidth}{!}{
    \begin{tabular}{c|c|c|c|c|c}
    \toprule[1pt]
        \multirow{2}{*}{\textbf{Method}} &
        \multirow{2}{*}{\textbf{Reference}} &
        \multicolumn{2}{c|}{\textbf{nuScenes $\rightarrow$ KITTI}} &
        \multicolumn{2}{c}{\textbf{Waymo $\rightarrow$ KITTI}} \\
        \cmidrule(lr){3-4} \cmidrule(lr){5-6} & &
        \textbf{$\textbf{AP}_{\textbf{BEV}} / \textbf{AP}_{\textbf{3D}}$} &
        \textbf{Closed Gap} &
        \textbf{$\textbf{AP}_{\textbf{BEV}} / \textbf{AP}_{\textbf{3D}}$} &
        \textbf{Closed Gap} \\
    \toprule
        Source & - & 45.50 / 13.78 & - & 65.64 / 26.74 & - \\
        Oracle & - & 83.29 / 73.45 & - & 83.29 / 73.45 & - \\
    \toprule
        Tent & ICLR'21 & 45.13 / 19.33 & $-$0.98\% / $+$9.30\%  & 68.54 / 34.57 & $+$16.43\% / $+$16.76\%  \\
        CoTTA & CVPR'22 & 68.81 / 47.61 & $+$61.68\% / $+$56.70\% & 67.46 / 35.34 & $+$10.31\% / $+$18.41\% \\
        Reg-TTA3D & ECCV'24 & 70.21 / 49.32 & $+$65.39\% / $+$59.56\% & 81.60 / 61.03 & $+$90.42\% / $+$73.41\% \\
        MOS & ICLR'25 & \underline{71.13} / \underline{51.11} & \underline{$+$67.82\%} / \underline{$+$62.56\%} & \underline{81.90} / \underline{64.16} & \underline{$+$92.12\%} / \underline{$+$80.11\%} \\
        DUO & ICCV'25 & 69.85 / 50.08 & $+$64.44\% / $+$60.83\% & 75.38 / 62.39 & $+$55.18\% / $+$76.32\% \\
        \rowcolor[HTML]{E6F1FF} Ours (\textbf{FIND}) & - & \textbf{75.21 / 54.66} & \textbf{$+$78.62\% / $+$68.51\%} & \textbf{82.35 / 67.09} & \textbf{$+$94.67\% / $+$86.38\%} \\
    \toprule[1pt]
    \end{tabular}
    }
    \label{tab:cross}
\end{table*}

\section{Experiments}
\label{sec:6}
\subsection{Experimental Setup}
For a fair comparison, we follow the identical evaluation pipeline with prior work~\cite{mos}, including baseline models, training configurations, and evaluation protocols.

\textbf{Datasets.} We conduct our experiments with three datasets (i.e., nuScenes~\cite{nuscenes}, Waymo~\cite{waymo}, KITTI~\cite{kitti}) to evaluate the robustness and generalization capability of our method in diverse cross-domain tasks: (1) \textit{Cross-Dataset:} nuScenes/Waymo$\rightarrow$KITTI introduces domain shifts arising from distinct locations and varying LiDAR beam configurations. (2) \textit{Corruptions:} KITTI-C~\cite{robo3d} simulates 8 corruption types including severe weather conditions, external disturbances and internal sensor failure to assess robustness against domain degradation. (3) \textit{Cross-Corruption:} Waymo$\rightarrow$KITTI-C presents a more rigorous domain shift, serving as a challenging evaluation benchmark.

\textbf{Compared Methods}. We compare our FIND with the following SOTA methods: CoTTA~\cite{cotta} employs a mean-teacher framework to provide supervision via augmentations and restores neurons to preserve knowledge. Reg-TTA3D~\cite{reg-tta} enhances box regression quality through confidence-guided refinement. MOS~\cite{mos} aggregates suitable historical checkpoints to integrate collective knowledge. DUO~\cite{duo} jointly minimizes semantic and geometric uncertainties in 3D object detection.

\textbf{Implementation Details.} Following the common setting in L3OD TTA, we evaluate our method and other baselines on SECOND-IoU~\cite{second}. We employ the same optimizer setting as MOS with a batch size of 8. For our FIND specifically, parameters $N$, $K$ and $\tau$ are assigned to default values of 12, 3, and 10, respectively.

\subsection{Main Results}

\textbf{Evaluation on Cross-Dataset.} We first compare our FIND with previous methods on two cross-dataset scenarios, as shown in Tab.~\ref{tab:cross}. The results yield three key observations. (1) When exposed to cross-location and cross-beam domain shifts, pretrained L3OD models suffer significant performance degradation. (2) Although existing TTA methods partially mitigate the adverse effects of domain-specific discrepancies, their efficacy remains suboptimal due to constraints arising from spatial entanglement. (3) Our method consistently outperforms compared approaches across all cross-dataset scenarios. Notably, the proposed FIND achieves the best performance, with average gains of \textbf{+2.27} $\text{AP}_{\text{BEV}}$ and \textbf{+3.24} $\text{AP}_{\text{3D}}$. Specifically, FIND closes more performance gap (\textbf{+10.80\% / +5.95\%} and \textbf{+2.55\% / +6.27\%}) on nuScenes$\rightarrow$KITTI and Waymo$\rightarrow$KITTI than prior SOTA method, respectively, approaching fully supervised performance.
\begin{table*}[t]
    \centering
    \caption{Comparisons with state-of-the-art methods on KITTI$\rightarrow$KITTI-C (heavy severity).  We report {$\text{AP}_{\text{3D}}$} of Car category in moderate difficulty. The complete results of moderate and easy severity levels are provided in Tab.~\ref{tab:corrupt-mid}-\ref{tab:corrupt-easy} of Appendix.~\ref{appendix:exp}.}
    \resizebox{\textwidth}{!}{
    \begin{tabular}{c|c|ccc|cc|ccc|c}
    \toprule[1pt]
        \multirow{2}{*}{\textbf{Method}} &
        \multirow{2}{*}{\textbf{Reference}} &
        \multicolumn{3}{c|}{\textbf{Weather}} &
        \multicolumn{2}{c|}{\textbf{Disturbance}} &
        \multicolumn{3}{c|}{\textbf{Sensor}} &
        \multirow{2}{*}{\textbf{Avg.}} \\ & &
        Fog & Rain & Snow &
        Motion. & Beam. &
        Cross T. & Inc. & Cross S.\\
    \toprule[0.5pt]
        Source & - & 67.72 & 75.99 & 59.32 & 38.77 & 53.85 & 75.21 & 27.49 & 41.24 & 54.95 \\    
        Tent & ICLR'21 & 68.84 & 76.22 & 59.69 & 38.35 & 53.76 & 75.10 & 28.49 & 41.35 & 55.23 \\
        CoTTA & CVPR'22 & 68.45 & 76.40 & 59.49 & 38.82 & 54.08 & 75.37 & 30.56 & 42.25 & 55.68 \\
        Reg-TTA3D & ECCV'24 & 68.00 & 77.51 & 60.72 & 40.34 & \underline{56.23} & 76.01 & 32.17 & \underline{44.02} & 56.88 \\
        MOS & ICLR'25 & 69.11 & 77.79 & \underline{62.72} & \underline{40.59} & 55.91 & 75.47 & \underline{34.53} & 43.68 & \underline{57.48} \\
        DUO & ICCV'25 & \underline{69.32} & \textbf{78.49} & 61.08 & 40.05 & 54.89 & \underline{76.27} & 33.50 & 42.10 & 56.96 \\
        \rowcolor[HTML]{E6F1FF} Ours (\textbf{FIND}) & - & \textbf{71.27} & \underline{78.23} & \textbf{64.89} & \textbf{42.96} & \textbf{58.04} & \textbf{76.88} & \textbf{38.49} & \textbf{46.20} & \textbf{59.62} \\
    \toprule[1pt]
    \end{tabular}
    }
    \label{tab:corrupt}
\end{table*}

\begin{table*}[t]
    \centering
    \caption{Comparisons with state-of-the-art methods on Waymo$\rightarrow$KITTI-C (heavy level).}
    \resizebox{\textwidth}{!}{
    \begin{tabular}{c|c|ccc|cc|ccc|c}
    \toprule[1pt]
        \multirow{2}{*}{\textbf{Method}} &
        \multirow{2}{*}{\textbf{Reference}} &
        \multicolumn{3}{c|}{\textbf{Weather}} &
        \multicolumn{2}{c|}{\textbf{Disturbance}} &
        \multicolumn{3}{c|}{\textbf{Sensor}} &
        \multirow{2}{*}{\textbf{Avg.}}\\ & &
        Fog & Rain & Snow &
        Motion. & Beam. &
        Cross T. & Inc. & Cross S. &\\
    \toprule[0.5pt]
        Source & - & 16.97 & 26.19 & 11.09 & 10.23 & 12.72 & 20.31 & 3.55 & 14.22 & 14.41 \\    
        Tent & ICLR'21 & 23.28 & 39.05 & 23.55 & 7.38 & 19.37 & 28.21 & 9.61 & 13.44 & 20.49 \\
        CoTTA & CVPR'22 & 31.39 & 50.10 & 25.11 & 18.99 & 27.00 & 35.19 & 10.10 & 13.26 & 26.39 \\
        Reg-TTA3D & ECCV'24 & 37.37 & \underline{59.16} & 27.90 & 23.16 & 32.13 & \textbf{41.35} & 13.80 & \underline{19.99} & 31.86 \\
        MOS & ICLR'25 & \textbf{43.26} & 58.32 & 28.87 & \underline{24.50} & 32.20 & 39.63 & \underline{14.96} & 19.19 & \underline{32.62} \\
        DUO & ICCV'25 & 39.59 & 52.88 & \textbf{30.33} & 23.74 & \underline{34.59} & 37.69 & 12.92 & 17.41 & 31.14 \\
        \rowcolor[HTML]{E6F1FF} Ours (\textbf{FIND}) & - & \underline{42.99} & \textbf{62.33} & \underline{30.10} & \textbf{24.66} & \textbf{35.02} & \underline{41.03} & \textbf{15.23} & \textbf{20.59} & \textbf{33.99} \\
    \toprule[1pt]
    \end{tabular}
    }
    \vspace{-0.5cm}
    \label{tab:cross-corrupt}
\end{table*}

\textbf{Evaluation on Corruptions.} We evaluate various methods on KITTI$\rightarrow$KITTI-C across 8 corruption types at hard level to validate the efficacy of the proposed FIND framework in mitigating distribution shifts. The experimental results, as reported in Tab.~\ref{tab:corrupt}, reveal several observations. (1) While existing methods yield only marginal improvements on the incomplete echoes (Inc.) corruption, our method achieves a substantial boost of \textbf{+11.00} $\text{AP}_{\text{3D}}$ compared with the pretrained model. This demonstrates that our frequency disentanglement approach effectively captures semantic content even under occlusion or sensor malfunction. (2) Our FIND method delivers significant performance gains, maintaining comparable or superior performance across all eight corruption types with an average improvement of \textbf{+2.14} $\text{AP}_{\text{3D}}$, further attesting to its robustness and effectiveness.

\begin{wraptable}{r}{0.5\textwidth}
    \centering
    \caption{Effects of components in our method. We conduct ablation studies of the invariance-driven alignment loss $\mathcal{L}_{\text{cons}}$, stability-guided regularization loss $\mathcal{L}_{\text{st}}$ and complementary constraint loss $\mathcal{L}_{\text{comp}}$ on cross-dataset scenarios. "Src." denotes the source model without adaptation.}
    \resizebox{\linewidth}{!}{
    \begin{tabular}{cccc|cc|cc}
    \toprule[1pt]
        \multirow{2}{*}{Src.} & \multirow{2}{*}{$\mathcal{L}_{\text{cons}}$} & \multirow{2}{*}{$\mathcal{L}_{\text{st}}$} & \multirow{2}{*}{$\mathcal{L}_{\text{comp}}$} & \multicolumn{2}{c|}{nuScenes$\rightarrow$KITTI} & \multicolumn{2}{c}{Waymo$\rightarrow$KITTI} \\
    \cmidrule(lr){5-6} \cmidrule(lr){7-8}
         & & & & $\textbf{AP}_{\textbf{BEV}}$ & $\textbf{AP}_{\textbf{3D}}$ & $\textbf{AP}_{\textbf{BEV}}$ & $\textbf{AP}_{\textbf{3D}}$ \\
    \toprule
        \ding{52} & & & & 45.50 & 13.78 & 65.64 & 26.74 \\
         & \ding{52} & & & 69.32 & 48.81 & 72.57 & 58.93 \\
         & & \ding{52} & & 60.84 & 35.87 & 69.88 & 43.91 \\
         & & \ding{52} & \ding{52} & 68.71 & 49.76 & 73.95 & 60.62 \\
         & \ding{52} & \ding{52} & & \underline{72.56} & \underline{52.49} & \underline{78.27} & \underline{64.48} \\
        \rowcolor[HTML]{E6F1FF} & \ding{52} & \ding{52} & \ding{52} & \textbf{75.21} & \textbf{54.66} & \textbf{82.35} & \textbf{67.09} \\
    \toprule[1pt]
    \end{tabular}
    }
    \label{tab:component}
\end{wraptable}

\textbf{Evaluation on Cross-Corruption.} We further conduct experiments on a composite shift scenario Waymo$\rightarrow$KITTI-C. As reported in Tab.~\ref{tab:cross-corrupt}, performance of detectors is significantly lower than that on Waymo$\rightarrow$KITTI in Tab.~\ref{tab:cross} and KITTI$\rightarrow$KITTI-C in Tab.~\ref{tab:corrupt}, attributed to the increased severity of the cross-domain shifts. In certain corruption scenarios, the adapted model even underperforms the non-adapted source model. Despite these challenges, FIND achieves a substantial gain of \textbf{+19.58} $\text{AP}_{\text{3D}}$ over the pretrained model.

\begin{wraptable}{r}{0.5\textwidth}
    \centering
    \caption{Computational efficiency on Waymo$\rightarrow$ KITTI-C using a single NVIDIA RTX 4090 with batch size 8. \textbf{Time} records the average processing time per point cloud in KITTI. \textbf{Memory} records max CUDA memory allocated during adaptation.}
    \label{tab:efficiency}
    \resizebox{\linewidth}{!}{%
    \begin{tabular}{l|cc}
        \toprule
        \textbf{Method} & \textbf{Time (ms)} & \textbf{Memory (MiB)} \\
        \midrule
        Tent & 88$\pm$5 & 7,476 \\
        CoTTA & 179$\pm$10 & 10,462 \\
        Reg-TTA3D & 206$\pm$14 & 8,564 \\
        MOS & 3088$\pm$160 & 14,134 \\
        DUO & 158$\pm$8 & 8,312 \\
        \rowcolor[HTML]{E6F1FF} Ours (\textbf{FIND}) & 194$\pm$18 & 11,540 \\
        \bottomrule
    \end{tabular}
    }
\end{wraptable}

\textbf{Computational Efficiency.} Having demonstrated the effectiveness of FIND across various domain shifts, we focus on the computational efficiency. Although FIND involves graph construction and Laplacian eigendecomposition, we use sparse matrix operations to reduce computational overhead (Appendix~\ref{appendix:spectral}) and parallelize the dual streams on the GPU. As shown in Tab.~\ref{tab:efficiency}, FIND requires substantially less processing time than MOS, the previous state-of-the-art, while remaining close to CoTTA. Appendix~\ref{appendix:efficiency} provides the detailed timing protocol and stage breakdown, which shows that spectral filtering accounts for a relatively small share of the total processing time. 
\vspace{-0.2cm}
\subsection{Ablation Study}
\label{sec:ablation}

In this subsection, without loss of generality, we conduct ablation studies on the cross-dataset scenario for the sake of brevity. Focusing on three pivotal components of FIND, \textit{Invariance-Driven Alignment}, \textit{Online Stability-Guided Regularization} and \textit{Complementary Constraint}, we perform various experiments to analyze their impacts.

\textbf{Effectiveness of Components.} We evaluate the contribution of individual components by comparing our full framework against variants that exclude key modules. As shown in Tab.~\ref{tab:component}, incorporating either $\mathcal{L}_{\text{cons}}$ or $\mathcal{L}_{\text{st}}$ enhances detection performance, yielding average gains of +15.4 / +33.6 and +9.8 / +19.6 $\text{AP}_{\text{BEV}}$ / $\text{AP}_{\text{3D}}$, respectively. Notably, $\mathcal{L}_{\text{comp}}$ enhances the adaptive capacity of $\mathcal{L}_{\text{st}}$ by preventing filters from collapsing into trivial solutions, highlighting the necessity of complementary constraint. These findings underscore the effectiveness of each component.

\begin{wrapfigure}{r}{0.5\textwidth}
    \vspace{-0.2cm}
    \centering
    \includegraphics[width=\linewidth]{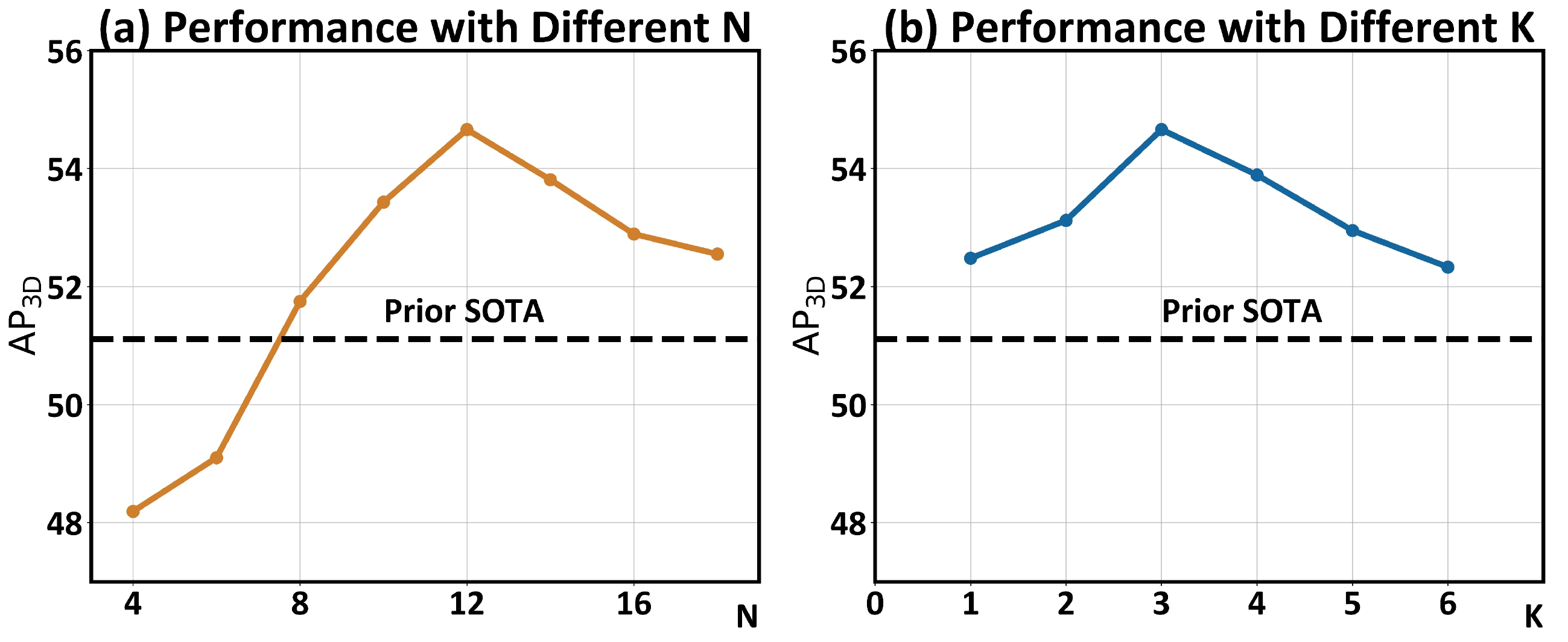}
    \caption{(a) Performance with varying numbers of bins $N$. (b) Performance with different B-Spline order $K$. Experiments are conducted on nuScenes$\rightarrow$KITTI.}
    \label{fig:ablation}
    \vspace{-0.3cm}
\end{wrapfigure}

\textbf{Hyperparameter Robustness.} We investigate the sensitivity of our B-Spline filter to the number of bins $N$ and the basis order $K$.
As shown in Fig.~\ref{fig:ablation}(a), a small $N$ limits the resolution for local spectral analysis, failing to isolate narrow-band invariant components.
In contrast, excessive $N$ induces increased susceptibility to certain bands, which destabilizes mask estimation and hinders disentanglement of domain-invariant features by overfitting to domain-specific variations.
Regarding $K$ as shown in Fig.~\ref{fig:ablation}(b), low-order fitting introduces derivative discontinuities that compromise the smoothness required for stable filtering, whereas excessively high degrees expand the local support range, causing spectral leakage between adjacent bands. The response curves in Appendix~\ref{appendix:ablation} can further serve as evidence.
Empirically, we adopt $N=12$ and $K=3$ as the optimal setting.

\vspace{-0.2cm}
\subsection{Disentanglement Visualization}

\begin{wrapfigure}{r}{0.5\textwidth}
    
    \centering
    \includegraphics[width=\linewidth]{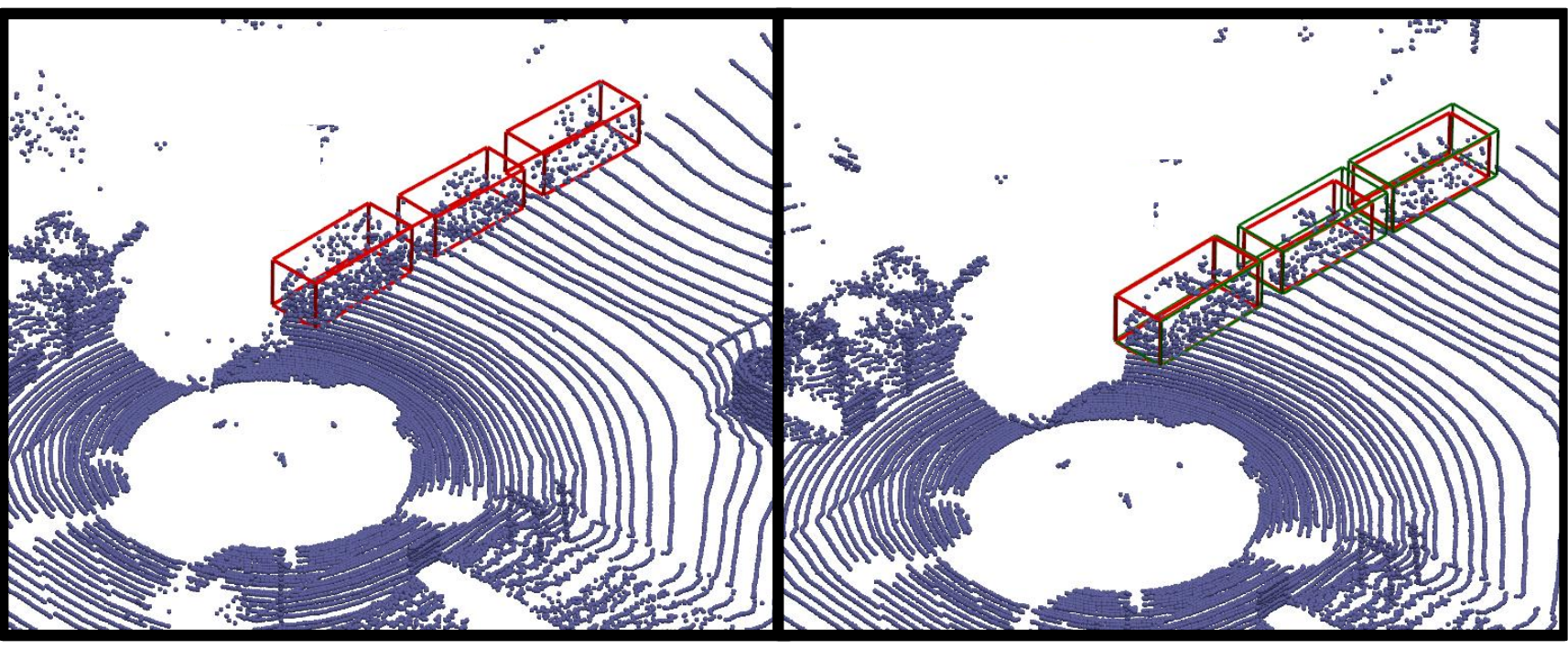}
    \caption{Visualization of point clouds in KITTI before and after filtering. \textit{Left:} Raw data entangled with sensor noise. \textit{Right:} Output from adaptive filters exhibiting geometric structure.}
    \label{fig:disentangle}
\end{wrapfigure}

Visualizations of the domain-invariant components, disentangled by adapted filters, demonstrate that our FIND framework guides the filters to learn the frequency distribution of invariant semantic content in the spectral domain. As illustrated in Figure~\ref{fig:disentangle}, this filtering mechanism extracts identifiable semantic content, enabling the model to yield reliable detection results.

\vspace{-0.2cm}
\section{Conclusion}
\vspace{-0.2cm}
In this paper, we propose FIND, a TTA framework that addresses the critical challenge of performance degradation in L3OD caused by the spatial entanglement of semantic content and domain style interference. Leveraging B-Spline fitting for locally adaptive spectral filters, FIND utilizes a dual-stream architecture to disentangle domain-invariant structure from domain-specific interference in frequency domain, providing a virtuous cycle between models and filters for adaptation. Extensive cross-dataset and corruption benchmarks demonstrate that FIND significantly outperforms SOTA methods by coupling model adaptation with filter optimization. Ultimately, we advance spectral graph analysis for non-Euclidean data, offering a robust pathway for reliable 3D perception. We hope that our work will motivate further investigation into frequency-domain perspectives on point cloud domain shifts and the development of advanced spectral disentangling techniques.

\textbf{Limitations.} First, although FIND effectively disentangles domain-invariant structure, the spectral transformation and B-Spline fitting processes introduce computational overhead during inference. Future work will investigate lightweight approximations of spectral components to enhance efficiency without compromising adaptation performance. Second, B-Spline efficacy relies on the initialization of knot vectors and control points. While our current setup is robust across tested datasets, significant variations in point cloud density may necessitate manual hyperparameter tuning. We aim to explore meta-adaptive mechanisms to automatically configure these parameters for diverse sensor resolutions.

\begin{ack}
This work was supported by the PKU-NTU Joint Research Institute (JRI) sponsored by a donation from the Ng Teng Fong Charitable Foundation, in part by AI Joint Lab of Future Urban Infrastructure sponsored by Fuzhou Chengtou New Infrastructure Group and Boyun Vision Co. Ltd.
\end{ack}

{
\bibliographystyle{abbrv}
\bibliography{neurips_2026}
}

\newpage
\appendix

\begin{center}
{\LARGE \textbf{FIND: Frequency Invariance Disentanglement for Test-Time Adaptation in LiDAR 3D  Detection\\ $\;$ \\ ————Appendix————}}
\end{center}

The structure of Appendix is as follows:
\begin{itemize}
    \item Appendix~\ref{appendix:theory} contains all missing theoretical basis in the main manuscript.
    \item Appendix~\ref{appendix:empirical} illustrates extended empirical analysis of spectral properties.
    \item Appendix~\ref{appendix:exp} presents further experimental results and computational efficiency analysis.
    \item Appendix~\ref{appendix:ablation} provides additional ablation studies on the robustness of FIND's adaptation process.
\end{itemize}

\section{Theoretical Basis Supplementation}
\label{appendix:theory}
Below, we provide detailed theoretical basis of the proposed FIND framework.

\subsection{B-Spline Fitting}
\label{appendix:b-spline}

In this subsection, we provide a comprehensive introduction to the B-spline theory utilized in our spectral filter design. We detail the recursive construction of basis functions, the calculation of derivatives for boundary constraints, the formulation of the linear system for coefficient determination, and the computational complexity of the fitting process.

\textbf{Knot Vector and Basis Definitions.}
B-splines (Basis splines) serve as a basis for spline functions within a specific domain. A B-spline of degree $K$ is defined over a sequence of non-decreasing real numbers known as the \textit{knot vector} $\Xi = \{ \xi_0, \xi_1, \dots, \xi_{M} \}$. The knots partition the parameter space into specific intervals, determining the continuity and shape of the basis functions.

For a spectral range defined on $[a, b]$ (in our method, $[0, 2]$) divided into $N$ bins, we employ a clamped uniform knot vector to ensure the curve passes through the endpoints. The knot vector is constructed as follows:
\begin{equation}
    \Xi = \{ \underbrace{a, \dots, a}_{K+1}, \xi_{K+1}, \dots, \xi_{K+N-1}, \underbrace{b, \dots, b}_{K+1} \},
\end{equation}
where the internal knots are uniformly spaced. The total number of knots is $M + 1 = N + 2K + 1$, supporting $(N+K)$ basis functions.

\textbf{Cox-de Boor Recursion Formula.}
The B-spline basis functions $B_{i,K}(x)$ are constructed recursively using the Cox-de Boor algorithm~\cite{bspline}. 

\textit{Base Case ($K=0$):}
The zero-degree basis functions are step functions, defined as:
\begin{equation}
    B_{i,0}(x) = 
    \begin{cases} 
    1 & \text{if } \xi_i \le x < \xi_{i+1}, \\
    0 & \text{otherwise}.
    \end{cases}
\end{equation}

\textit{Recursive Step ($K > 0$):}
For degree $K$, the basis function $B_{i,K}(x)$ is a linear combination of two basis functions of degree $K-1$:
\begin{equation}
    B_{i,K}(x) = \frac{x - \xi_i}{\xi_{i+K} - \xi_i} B_{i,K-1}(x) + \frac{\xi_{i+K+1} - x}{\xi_{i+K+1} - \xi_{i+1}} B_{i+1,K-1}(x).
\end{equation}
By convention, if a denominator is zero, the corresponding term is treated as zero. This recursion ensures that $B_{i,K}(x)$ has \textit{compact local support property}, being non-zero only over the interval $[\xi_i, \xi_{i+K+1})$. This property is critical for the "local control" mentioned in Sec.~\ref{sec:3}, as changing a coefficient $c_i$ only affects the curve locally.

\textbf{Derivatives of B-Splines.}
To enforce the vanishing derivative constraints at the boundaries (as described in Sec.~\ref{sec:5.1} via matrix $\mathbf{D}$), we require the analytical derivatives of the B-spline functions. The derivative of a B-spline of degree $K$ with respect to $x$ is given by:
\begin{equation}
    \frac{d}{dx} B_{i,K}(x) = K \left( \frac{B_{i,K-1}(x)}{\xi_{i+K} - \xi_i} - \frac{B_{i+1,K-1}(x)}{\xi_{i+K+1} - \xi_{i+1}} \right).
\end{equation}
Higher-order derivatives can be computed by applying this formula recursively. Generally, the $j$-th derivative $B^{(j)}_{i,K}(x)$ reduces to a linear combination of B-splines of degree $K-j$. This allows us to construct the derivative constraint matrix $\mathbf{D}$ precisely without numerical approximation errors.

\textbf{Properties.}
The B-Spline basis functions exhibit several critical properties for numerical stability and local control:
\begin{itemize}
    \item \textbf{Compact Local Support:} $B_{i,K}(x) > 0$ only for $x \in (\xi_i, \xi_{i+K+1})$. This ensures that changing a control coefficient $c_i$ affects the curve only locally.
    \item \textbf{Partition of Unity:} For any $x$ within the valid domain, $\sum_i B_{i,K}(x) = 1$.
    \item \textbf{Positivity:} $B_{i,K}(x) \geq 0$ for all $x$.
\end{itemize}

\textbf{Solving for Control Coefficients.}
As defined in Eq.~(\ref{equ:filter}), the spectral response is $h(\lambda) = \sum_{i=0}^{N+K-1} c_i B_{i,K}(\lambda)$. We aim to determine the coefficient vector $\mathbf{C} = [c_0, \dots, c_{N+K-1}]^T\in\mathbb{R}^{N+K}$ subject to two sets of constraints.

\begin{enumerate}
    \item \textbf{Interpolation Constraints:} We require the filter to match the learnable response $\delta_j$ at $N+1$ knot locations $\lambda_j$. This yields $N+1$ linear equations:
    \begin{equation}
        \sum_{i=0}^{N+K-1} c_i B_{i,K}(\lambda_j) = \delta_j, \quad j = 0, \dots, N.
    \end{equation}
    In matrix form, this is expressed as $\mathbf{B}^T \mathbf{C} = \mathbf{\Delta}^T$, where $\mathbf{B}_{ij} = B_{i,K}(\lambda_j)$ and $\mathbf{\Delta} = [\delta_0, \dots, \delta_N]$.
    
    \item \textbf{Boundary Derivative Constraints:} To ensure smoothness and satisfy boundary conditions (i.e., zero response slope at the start of the spectrum), we constrain the first $K-1$ derivatives to be zero at $\lambda_0$. For the $j$-th order derivative ($1 \le j \le K-1$):
    \begin{equation}
        h^{(j)}(\lambda_0) = \sum_{i=0}^{N+K-1} c_i B_{i,K}^{(j)}(\lambda_0) = 0.
    \end{equation}
    This forms the system $\mathbf{D}^T \mathbf{C} = \mathbf{0}$, where $\mathbf{D}_{ij} = B_{i,K}^{(j+1)}(\lambda_0)$ (assuming $0$-indexed columns for derivatives $1 \dots K-1$).
\end{enumerate}

Combining the interpolation and derivative constraints, we form a square system of linear equations. The total number of constraints is $(N+1) + (K-1) = N+K$, which matches the number of unknown coefficients in $\mathbf{C}$. 

We define the unified constraint matrix $\mathbf{M} \in \mathbb{R}^{(N+K) \times (N+K)}$ by concatenating $\mathbf{B}$ and $\mathbf{D}$:
\begin{equation}
    \mathbf{M} = [\mathbf{B}, \mathbf{D}].
\end{equation}
The full system is written as:
\begin{equation}
    \mathbf{M}^T \mathbf{C} = \begin{bmatrix} \mathbf{\Delta}^T \\ \mathbf{0}_{K-1}^T \end{bmatrix}.
\end{equation}
Solving for $\mathbf{C}$ yields the solution provided in Eq. (\ref{equ:sol}):
\begin{equation}
    \mathbf{C} = (\mathbf{M}^T)^{-1} [\mathbf{\Delta}, \mathbf{0}_{K-1}]^T = [\mathbf{B}, \mathbf{D}]^{-T} [\mathbf{\Delta}, \mathbf{0}_{K-1}]^T.
\end{equation}

\textbf{Complexity Analysis.}
The computational complexity of the B-spline filter design consists of two parts: evaluation (i.e., inference in filtering modules) and optimization (i.e., solving coefficients during adaptation).

\begin{enumerate}
    \item \textbf{Inference Complexity:} Calculating a value $h(\lambda)$ at a single point requires evaluating the non-zero basis functions. Due to compact support, at any point $\lambda \in [\xi_i, \xi_{i+1})$, at most $K+1$ basis functions (specifically $B_{i-K, K}, \dots, B_{i, K}$) are non-zero. Using the Cox-de Boor algorithm, computing a value of degree $K$ requires triangular computation. Totally, the inference of a single point takes $O(K^2)$ operations.

    \item \textbf{Coefficient Solution Complexity:} Solving for $\mathbf{C}$ requires inverting the linear system in Eq.~(\ref{equ:sol}), where $[\mathbf{B}, \mathbf{D}]^T$ is a square matrix with size of $(N+K) \times (N+K)$.
    \begin{itemize}
        \item \textbf{Naive Approach:} Standard Gaussian elimination or matrix inversion would yield a complexity of $O((N+K)^3)$, which is computationally expensive for high-resolution spectral bins (large $N$).
        \item \textbf{Banded Matrix Approach:} Due to the compact support property, the matrix $\mathbf{M}$ is \textit{banded}. The entry $\mathbf{M}_{ji} = B_{i,K}(\lambda_j)$ is non-zero only if $\lambda_j$ falls within the support of $B_{i,K}$, i.e., $\xi_i \le \lambda_j < \xi_{i+K+1}$. This implies that for a specific row $j$, non-zero entries only exist for columns $i$ typically in the range $[j-K, j]$. Consequently, the bandwidth $w$ of matrix $\mathbf{M}$ is $K$.
    \end{itemize}
    Solving a linear system for a banded matrix with size $M$ and bandwidth $w$ using LU decomposition requires approximately $O(M \cdot w^2)$ operations, rather than $O(M^3)$.

    Substituting our parameters:
    \begin{equation}
        T_{solve} \approx O((N+K) \cdot K^2).
    \end{equation}

\end{enumerate}

\subsection{Spectral Consistency Across Samples}
\label{appendix:spectral}

Since the GFT basis is derived from the Laplacian of each local graph, the eigenvectors differ across samples. A natural question is whether spectral comparison across different point clouds remains meaningful. We discuss this from three complementary perspectives.

\textbf{Universal spectral semantics via normalized Laplacian.}
The symmetric normalized Laplacian $L_{\text{sym}} = I - D^{-1/2}AD^{-1/2}$ has eigenvalues universally bounded in $[0, 2]$, regardless of graph size, density, or connectivity~\cite{shuman2013emerging}. The eigenvalue magnitude carries a consistent geometric interpretation: a small $\lambda$ corresponds to a signal that varies smoothly over the graph (capturing large-scale structure), while a large $\lambda$ corresponds to a signal that oscillates rapidly between neighboring nodes (capturing local details and edges)~\cite{shuman2013emerging,hu2021graph}. This smoothness-based semantic interpretation is intrinsic to the Laplacian operator and does not require sharing exact eigenvectors across graphs.

This principle underpins a broad family of spectral graph methods. Prior work~\cite{levie2021transferability} has formally shown that spectral filters defined as functions of the Laplacian eigenvalues can transfer across graphs with different topologies, because such filters operate on the eigenvalue axis rather than on specific eigenvectors. Similarly, spectral methods for 3D point cloud analysis~\cite{hu2022exploring,liu2023point} and cross-graph matching~\cite{bastico2024coupled} exploit this property to define meaningful operations across varying graph structures.

\textbf{Controlled graph construction.}
In FIND, we further promote spectral comparability through controlled graph construction. Graphs are built on object-level point clouds within predicted bounding boxes rather than on entire scenes, yielding local graphs of similar scale ($O(10^2)$ nodes). 
The graph is a sparse kNN graph with multiple scattered connected blocks, providing constant-level time optimization for subsequent eigenvalue decomposition. Therefore, although the worst-case complexities of graph construction and eigen-decomposition are $O(n^2)$ and $O(n^3)$, respectively, the practical overhead remains moderate.

Furthermore, the kNN graph with Gaussian kernel weights (Sec.~\ref{sec:3}) ensures that the edge structure reflects local geometric proximity. These design choices make the graph topology primarily dependent on the object's intrinsic geometry rather than scene-level layout, promoting consistent spectral distributions for objects of similar categories.

\textbf{B-Spline filters absorb residual misalignment.}
Even after the above measures, minor sample-level spectral shifts may persist due to variations in point density or partial occlusion. Our B-Spline filter design (Eq.~\ref{equ:filter}) addresses this through its compact local support property: each basis function $B_{i,K}(\lambda)$ is active only within a localized spectral interval, so the filter response at each frequency region is determined by a small number of neighboring control points. This means the filter operates on spectral \emph{regions} rather than on individual eigenvalues, naturally absorbing minor misalignments and ensuring regionally consistent spectral processing across different samples.

\section{Extended Empirical Analysis}
\label{appendix:empirical}

This section provides extended empirical analysis to complement the observations in Sec.~\ref{sec:4}. We first present controlled sensor-only experiments that isolate the effect of sensor/weather conditions from scene content (Sec.~\ref{appendix:empirical:controlled}), and then demonstrate the generality of our findings across multiple domain shift scenarios (Sec.~\ref{appendix:empirical:cross}).

\subsection{Controlled Sensor-Only Experiments (KITTI→KITTI-C)}
\label{appendix:empirical:controlled}

A potential confound in the main-text analysis (Fig.~\ref{fig:obser}, nuScenes$\rightarrow$KITTI) is that spectral differences could arise from scene content variation rather than sensor-induced domain shift. To disentangle these factors, we conduct controlled experiments on KITTI$\rightarrow$KITTI-C~\cite{3drobust}, where various sensor/weather corruptions (fog, rain, beam dropout, crosstalk, etc.) are applied to the \emph{same} KITTI scenes, keeping the underlying geometry fixed while only varying sensor conditions.

\begin{figure}[h]
    \centering
    \includegraphics[width=\linewidth]{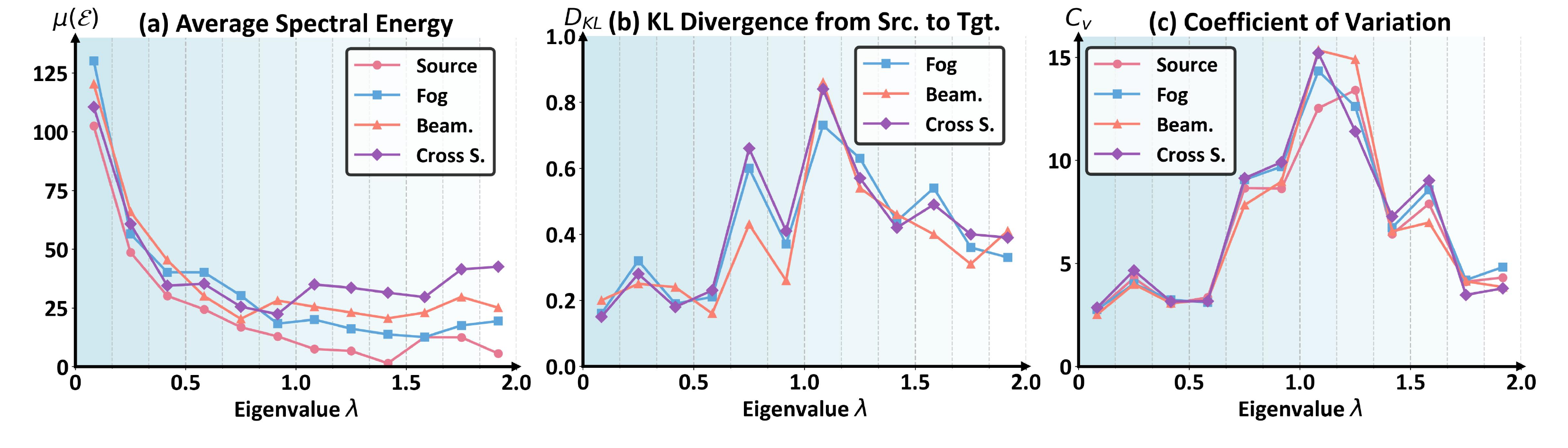}
    \caption{Controlled sensor-only empirical analysis on KITTI$\rightarrow$KITTI-C. The underlying scene geometry is fixed while only sensor/weather conditions vary. \textit{(a)} Average spectral energy $\mu(\mathcal{E})$ span all bands regardless of corruption type. \textit{(b)} Per-band $D_{KL}$ fluctuation pattern is consistent under different corruptions. \textit{(c)} $C_v$ profiles remain stable across different corruption types.}
    \label{fig:empirical_controlled}
\end{figure}

\textbf{Supporting Observation~1 — cross-band entanglement.}
As shown in Fig.~\ref{fig:empirical_controlled}(a), despite identical scene geometry, spectral energy changes are observed across all frequency bands for every corruption type. No single frequency band is immune to style perturbations, confirming that domain-specific interference is entangled with semantic content across the entire spectrum and is not confined to any particular frequency range.

\textbf{Supporting Observation~2 — narrow-band $D_{KL}$ fluctuations.}
Fig.~\ref{fig:empirical_controlled}(b) displays the per-band KL divergence $D_{KL}^{(i)}$ from clean KITTI to each corruption type. The sharp, narrow-band fluctuation pattern observed in the cross-dataset setting (Fig.~\ref{fig:obser}(b), nuScenes$\rightarrow$KITTI) is reproduced under these controlled sensor-only conditions, confirming that the fluctuations reflect genuine frequency-specific sensitivity to environmental variation rather than artifacts of differing scene composition.

\textbf{Supporting Observation~3 — consistency of $C_v$ profiles.}
Fig.~\ref{fig:empirical_controlled}(c) overlays the per-band $C_v$ curves computed within KITTI-C under different corruption types. The instability profiles are highly consistent across corruption scenarios, indicating that $C_v$ captures an intrinsic property of each frequency band's susceptibility to environmental variation, independent of the specific corruption applied.

\subsection{Cross-Scenario Generalization}
\label{appendix:empirical:cross}

To verify that the spectral properties in Sec.~\ref{sec:4} are not specific to the nuScenes$\rightarrow$KITTI scenario, we extend the analysis to Waymo$\rightarrow$KITTI and KITTI$\rightarrow$KITTI-C.

\begin{figure}[h]
    \centering
    \includegraphics[width=\linewidth]{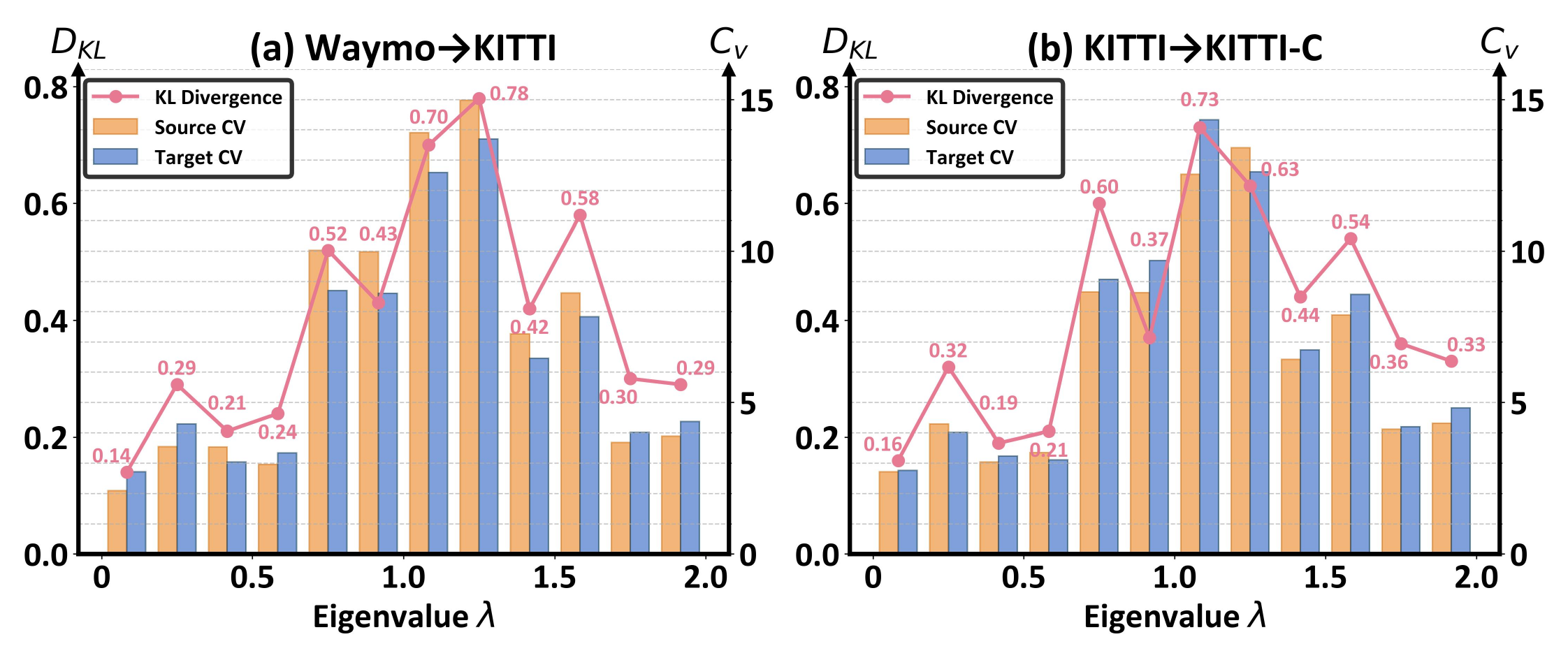}
    \caption{Empirical analysis of KL Divergence and Coefficient of Variation on Waymo→KITTI and KITTI→KITTI-C (fog, heavy) scenarios.}
    \label{fig:empirical_scenarios}
\end{figure}

\textbf{$\mathbf{D_{KL}}$ fluctuation across scenarios.}
The line graphs in Fig.~\ref{fig:empirical_scenarios} show the per-band $D_{KL}$ for Waymo$\rightarrow$KITTI and KITTI$\rightarrow$KITTI-C (fog corruption), respectively. Both scenarios exhibit the same characteristic pattern observed in Fig.~\ref{fig:obser}(b): $D_{KL}$ persists across the full spectrum with sharp band-to-band fluctuations, supporting that this phenomenon is a general property of spectral domain shift rather than a dataset-specific artifact.

\textbf{$\mathbf{C_v-D_{KL}}$ correlation across scenarios.}
Fig.~\ref{fig:empirical_scenarios} also presents the $C_v$ vs.\ $D_{KL}$ scatter plots for each scenario.
Tab.~\ref{tab:spearman} reports the Spearman rank correlation coefficients. In all cases, $C_v$ and $D_{KL}$ exhibit strong monotonic correlation ($\rho \ge 0.94$, $p < 10^{-5}$) for both source and target domains, confirming the generality and statistical robustness of the $C_v$--$D_{KL}$ relationship.

\begin{table}[h]
    \centering
    \caption{Spearman rank correlation ($\rho$) between $C_v$ and $D_{KL}$ across scenarios. All correlations are statistically significant ($p < 10^{-5}$).}
    \label{tab:spearman}
    \begin{tabular}{lcccc}
        \toprule
        \multirow{2}{*}{Scenario} & \multicolumn{2}{c}{Source Domain} & \multicolumn{2}{c}{Target Domain} \\
        \cmidrule(lr){2-3} \cmidrule(lr){4-5}
        & $\rho$ & $p$-value & $\rho$ & $p$-value \\
        \midrule
        nuScenes $\rightarrow$ KITTI   & 0.97 & $1.75 \times 10^{-7}$ & 0.97 & $1.75 \times 10^{-7}$ \\
        Waymo $\rightarrow$ KITTI      & 0.96 & $7.80 \times 10^{-7}$ & 0.96 & $1.18 \times 10^{-6}$ \\
        KITTI $\rightarrow$ KITTI-C (fog) & 0.95 & $2.04 \times 10^{-6}$ & 0.94 & $3.93 \times 10^{-6}$ \\
        \bottomrule
    \end{tabular}
\end{table}

\section{Further Experiments}
\label{appendix:exp}
In this section, we extend our evaluation of FIND beyond the main experimental settings. By varying the target conditions and comparing alternative approaches, we assess the consistency of its performance gains alongside the computational cost of adaptation. These experiments allow us to thoroughly examine the robustness of our approach against different types of domain shifts.

\subsection{Distinct Object Categories}
We compare our FIND with state-of-the-art methods on three cross-dataset scenarios covering Car, Pedestrian, and Cyclist categories, as shown in Tab.~\ref{tab:all-cross}. The results yield three key observations. (1) When transferring between domains with distinct sensor configurations and environments, source models suffer from severe performance degradation on all categories of objects. (2) Although existing methods partially mitigate domain shifts, their efficacy varies across categories, leaving room for improvement in multi-class adaptation. (3) Our method consistently outperforms compared approaches in the majority of scenarios. Notably, the proposed FIND achieves the best performance in 7 out of 9 metrics. Specifically, FIND dominates the Car category across all scenarios, surpassing the second-best method MOS by \textbf{+3.55}, \textbf{+2.93}, and \textbf{+0.91} $\text{AP}_{\text{3D}}$ on nuScenes$\rightarrow$KITTI, Waymo$\rightarrow$KITTI, and Waymo$\rightarrow$nuScenes, respectively, demonstrating robust generalization capability.

\begin{table*}[b]
    \centering
    \caption{Comparisons with state-of-the-art methods on nuScenes$\rightarrow$KITTI, Waymo$\rightarrow$KITTI and Waymo$\rightarrow$nuScenes scenarios. We record $\text{AP}_{\text{3D}}$ of Car, Pedestrian (Ped.), Cyclist (Cyc.) categories in moderate difficulty. We highlight the best and second results with \textbf{bold} and \underline{underline} respectively.}
    \resizebox{0.9\textwidth}{!}{
    \begin{tabular}{c|c|ccc|ccc|ccc}
    \toprule[1pt]
        \multirow{2}{*}{\textbf{Method}} &
        \multirow{2}{*}{\textbf{Reference}} &
        \multicolumn{3}{c|}{\textbf{nuScenes $\rightarrow$ KITTI}} &
        \multicolumn{3}{c|}{\textbf{Waymo $\rightarrow$ KITTI}} & \multicolumn{3}{c}{\textbf{Waymo $\rightarrow$ nuScenes}} \\
        \cmidrule(lr){3-5} \cmidrule(lr){6-8} \cmidrule(lr){9-11} & &
        Car & Ped. & Cyc. & Car & Ped. & Cyc. & Car & Ped. & Cyc. \\
    \toprule
        Source & - & 13.78 & 27.75 & 8.37 & 26.74 & 36.01 & 38.29 & 17.22 & 0.78 & 1.03 \\
        Oracle & - & 73.45 & 41.33 & 60.32 & 73.45 & 41.33 & 60.32 & 34.87 & 18.92 & 11.73 \\
    \toprule
        Tent & ICLR'21 & 19.33 & 29.55 & 11.12 & 34.57 & 38.39 & 41.56 & 17.83 & 1.78 & 0.92 \\
        CoTTA & CVPR'22 & 47.61 & 33.45 & 12.46 & 35.34 & 42.05 & 44.75 & 18.12 & 1.73 & 2.78 \\
        Reg-TTA3D & ECCV'24 & 49.32 & 33.63 & 16.44 & 61.03 & 45.53 & 49.77 & 18.26 & \textbf{2.23} & \textbf{4.73} \\
        MOS & ICLR'25 & \underline{51.11} & \underline{38.45} & \underline{18.16} & \underline{64.16} & \underline{46.51} & \underline{51.36} & \underline{18.75} & 1.79 & \underline{4.51} \\
        DUO & ICCV'25 & 50.08 & 37.51 & 17.49 & 62.39 & 45.11 & 49.63 & 18.63 & 0.94 & 3.90 \\
        \rowcolor[HTML]{E6F1FF} Ours (\textbf{FIND}) & - & \textbf{54.66} & \textbf{39.43} & \textbf{19.31} & \textbf{67.09} & \textbf{47.92} & \textbf{52.22} & \textbf{19.66} & \underline{1.99} & 4.35 \\
    \toprule[1pt]
    \end{tabular}
    }
    \label{tab:all-cross}
\end{table*}

\subsection{Different Severity Level Corruption}
We further provide more discussions surrounding various severity data corruptions (i.e., moderate and easy) based on the results shown in Tab.~\ref{tab:corrupt-mid}-\ref{tab:corrupt-easy}, which clearly gives additional observations: 
1) With the escalation of severity level, the source models suffer a larger performance decline within various corruptions. Pretrained models achieve obvious performance drop from easy level to heavy level, which significantly heightens the challenge for test-time adaptation.
2) Existing TTA methods struggle to recover performance under different levels of corruptions, highlighting the limitations of conventional approaches that focus exclusively on the spatial domain.
3) Despite these challenges, our FIND framework consistently achieves the best average performance across all corruption types. This robust performance demonstrates that our frequency disentanglement framework effectively stabilizes spatial predictions, providing reliable adaptation even under different shift levels.

\begin{table*}[t]
    \centering
    \caption{Comparisons with state-of-the-art methods on KITTI$\rightarrow$KITTI-C (moderate severity). We record $\text{AP}_{\text{3D}}$ of Car category in moderate difficulty and highlight the best and second results with \textbf{bold} and \underline{underline} respectively.}
    \resizebox{0.9\textwidth}{!}{
    \begin{tabular}{c|c|ccc|cc|ccc|c}
    \toprule[1pt]
    
    \multirow{2}{*}{\textbf{Method}} &
    \multirow{2}{*}{\textbf{Reference}} &
    \multicolumn{3}{c|}{\textbf{Weather}} &
    \multicolumn{2}{c|}{\textbf{Disturbance}} &
    \multicolumn{3}{c|}{\textbf{Sensor}} &
    \multirow{2}{*}{\textbf{Avg.}}\\ & &
    Fog & Rain & Snow &
    Motion. & Beam. &
    Cross T. & Inc. & Cross S.\\
    
    \toprule[0.5pt]
    
    Source & - & 72.36 & 79.13 & 64.09 & 39.51 & 57.53 & 78.26 & 30.39 & 47.21 & 58.56 \\    
    TENT & ICLR'21 & 73.07 & 78.98 & 64.18 & 39.66 & 58.01 & 78.19 & 30.50 & 47.34 & 58.74 \\
    CoTTA & CVPR'22 & 73.11 & 80.59 & 65.93 & 40.51 & 59.09 & 78.72 & 35.34 & 48.36 & 60.21 \\
    REG-TTA3D & ECCV'24 & 73.59 & \underline{81.93} & 67.67 & 41.37 & 60.26 & 79.06 & 39.51 & \underline{49.40} & 61.60 \\
    MOS & ICLR'25 & \underline{74.02} & 81.91 & \underline{67.87} & \underline{41.63} & \underline{60.48} & 79.10 & \underline{40.29} & 49.28 & \underline{61.82} \\
    DUO & ICCV'25 & 73.75 & 81.59 & 67.19 & 41.21 & 59.95 & \textbf{79.47} & 38.65 & 49.03 & 61.36 \\
    \rowcolor[HTML]{E6F1FF} Ours (\textbf{FIND}) & - & \textbf{74.53} & \textbf{82.07} & \textbf{69.39} & \textbf{42.82} & \textbf{60.77} & \underline{79.27} & \textbf{43.19} & \textbf{49.51} & \textbf{62.69} \\
    
    \toprule[1pt]
    \end{tabular}
    }
    \label{tab:corrupt-mid}
\end{table*}

\begin{table*}[t]
    \centering
    \caption{Comparisons with state-of-the-art methods on KITTI$\rightarrow$KITTI-C (easy severity). We record $\text{AP}_{\text{3D}}$ of Car category in moderate difficulty and highlight the best and second results with \textbf{bold} and \underline{underline} respectively.}
    \resizebox{0.9\textwidth}{!}{
    \begin{tabular}{c|c|ccc|cc|ccc|c}
    \toprule[1pt]
    
    \multirow{2}{*}{\textbf{Method}} &
    \multirow{2}{*}{\textbf{Reference}} &
    \multicolumn{3}{c|}{\textbf{Weather}} &
    \multicolumn{2}{c|}{\textbf{Disturbance}} &
    \multicolumn{3}{c|}{\textbf{Sensor}} &
    \multirow{2}{*}{\textbf{Avg.}}\\ & &
    Fog & Rain & Snow &
    Motion. & Beam. &
    Cross T. & Inc. & Cross S.\\
    
    \toprule[0.5pt]
    
    Source & - & 84.51 & 88.37 & 71.29 & 43.49 & 75.38 & 88.10 & 48.11 & 68.02 & 70.91 \\    
    TENT & ICLR'21 & 84.21 & 88.68 & 71.54 & 43.88 & 76.05 & 88.30 & 50.16 & 69.06 & 71.49 \\
    CoTTA & CVPR'22 & 84.89 & 88.54 & 73.53 & 44.46 & 76.98 & 88.70 & 53.66 & 69.88 & 72.58 \\
    REG-TTA3D & ECCV'24 & 85.03 & 88.91 & 74.79 & 44.58 & 78.27 & 88.90 & 58.39 & 70.68 & 73.69 \\
    MOS & ICLR'25 & \underline{85.22} & 89.32 & 75.07 & 44.70 & \underline{78.74} & 89.22 & \underline{59.13} & \underline{71.87} & \underline{74.16} \\
    DUO & ICCV'25 & \textbf{85.29} & \underline{89.36} & \underline{75.27} & \underline{44.74} & 78.52 & \textbf{89.72} & 58.44 & 71.32 & 74.08 \\
    \rowcolor[HTML]{E6F1FF} Ours (\textbf{FIND}) & - & \textbf{85.29} & \textbf{89.42} & \textbf{75.86} & \textbf{44.83} & \textbf{79.56} & \underline{89.23} & \textbf{60.75} & \textbf{73.52} & \textbf{74.81} \\
    
    \toprule[1pt]
    \end{tabular}
    }
    \label{tab:corrupt-easy}
\end{table*}

\subsection{Computational Efficiency}
\label{appendix:efficiency}

We further analyze the computational cost of FIND to identify the main sources of runtime overhead. Table~\ref{tab:efficiency-stages} separates FIND's processing time into four mutually exclusive stages. Bounding-box prediction supplies the object regions used for graph construction. The spectral filtering stage includes Laplacian eigendecomposition, GFT, evaluation of $h(\lambda)$, filtering for both streams, and inverse GFT. The adaptation stage updates the detector and learnable filter responses using the consistency, stability, and complementarity objectives. Spectral filtering accounts for 13.92\% of the total processing time, while dual-stream adaptation accounts for 45.36\%.

\begin{table}[t]
    \centering
    \caption{Processing-time breakdown of FIND under the same measurement setting. Each stage's time is normalized by the number of processed frames.}
    \label{tab:efficiency-stages}
    \resizebox{0.65\textwidth}{!}{
    \begin{tabular}{l|c|c}
        \toprule[1pt]
        \textbf{Stage} & \textbf{Time (ms)} & \textbf{Proportion} \\
        \toprule[0.5pt]
        Bounding-box prediction & 73 & 37.63\% \\
        Semantic-aware graph construction & 6 & 3.09\% \\
        B-Spline spectral filtering & 27 & 13.92\% \\
        Dual-stream disentangled adaptation & 88 & 45.36\% \\
        \toprule[0.5pt]
        \rowcolor[HTML]{E6F1FF} Full \textbf{FIND} & 194 & 100.00\% \\
        \toprule[1pt]
    \end{tabular}
    }
\end{table}

\subsection{Comparison with Denoising and Frequency-Domain Adaptation Methods}
\label{appendix:additional-comparisons}

To complement the filter ablations in Tab.~\ref{tab:filter}, we further compare FIND with the graph-wavelet point-cloud denoising method SC-NNK~\cite{watanabe2023scnnk} and frequency-domain TTA methods L-TTA~\cite{shin2024ltta} and SPA~\cite{niu2025spa}. We use the same SECOND-IoU detector and evaluation settings as in Sec.~\ref{sec:6}. This comparison evaluates alternative denoising and adaptation approaches, while the filter ablations examine the choice of filtering mechanism within FIND.

As shown in Tab.~\ref{tab:additional-comparisons}, FIND outperforms all three methods across the tested scenarios. The relative performance of the alternatives varies with the type of domain shift: SPA performs best among them on the two cross-dataset tasks, whereas L-TTA performs best on KITTI$\rightarrow$KITTI-C. FIND maintains its advantage in both settings, indicating that its benefits extend across sensor and scene differences as well as data corruptions. Together with the filter ablations, these results support coupling locally adaptive spectral filtering with detector adaptation for LiDAR 3D detection.

\begin{table}[t]
    \centering
    \caption{Comparisons with point-cloud denoising and frequency-domain TTA methods. We report $\text{AP}_{\text{3D}}$ of the Car category at moderate detection difficulty. KITTI$\rightarrow$KITTI-C results are averaged over eight corruption types at heavy severity. We highlight the best and second results with \textbf{bold} and \underline{underline}, respectively.}
    \label{tab:additional-comparisons}
    \resizebox{0.9\textwidth}{!}{
    \begin{tabular}{c|c|ccc}
        \toprule[1pt]
        \textbf{Method} & \textbf{Reference} & \textbf{nuScenes $\rightarrow$ KITTI} & \textbf{Waymo $\rightarrow$ KITTI} & \textbf{KITTI $\rightarrow$ KITTI-C} \\
        \toprule[0.5pt]
        Source & - & 13.78 & 26.74 & 54.95 \\
        SC-NNK & ICASSP'23 & 42.69 & 52.37 & 55.46 \\
        L-TTA & NeurIPS'24 & 44.57 & 51.33 & \underline{56.25} \\
        SPA & ICML'25 & \underline{48.19} & \underline{57.56} & 55.92 \\
        \rowcolor[HTML]{E6F1FF} Ours (\textbf{FIND}) & - & \textbf{54.66} & \textbf{67.09} & \textbf{59.62} \\
        \toprule[1pt]
    \end{tabular}
    }
\end{table}

\subsection{Continual Adaptation under Abrupt Domain Shifts}
\label{appendix:continual}

We further evaluate whether FIND can maintain adaptation when the target distribution changes during deployment. We construct a continual Fog$\rightarrow$Rain$\rightarrow$Snow stream on KITTI$\rightarrow$KITTI-C and compare FIND with MOS. Both methods are initialized once and process the same stream without resets, domain-boundary labels, or corruption labels. Tab.~\ref{tab:continual} compares this setting with independent adaptation on each corruption.

In FIND, the EMA with $\beta=0.001$ smooths only the point-selection threshold $\eta_{th}$, which tracks the mean normalized invariant-retention ratio of selected points. As the invariant filter improves, threshold calibration helps the mask retain domain-invariant geometry and remove points dominated by domain-specific components. The consistency objective is computed and backpropagated at every target batch, so the detector and B-Spline filters continue to receive per-batch updates.

The results show that FIND retains performance close to independent adaptation throughout the tested stream. Its average $\text{AP}_{\text{3D}}$ decreases by 0.37 points under continual adaptation, compared with 1.04 points for MOS. The small difference between FIND's two protocols indicates that threshold smoothing is compatible with sustained adaptation across these abrupt shifts, while the detector and filters continue to respond to incoming target batches.

\begin{table}[t]
    \centering
    \caption{Continual adaptation under Fog$\rightarrow$Rain$\rightarrow$Snow shifts on KITTI$\rightarrow$KITTI-C at heavy severity. We report $\text{AP}_{\text{3D}}$ of the Car category at moderate detection difficulty. ``Independent'' denotes separate adaptation to each corruption, while ``Continual'' denotes adaptation along the stream without resets or domain labels. We highlight the best results within each adaptation protocol with \textbf{bold}.}
    \label{tab:continual}
    \resizebox{0.75\textwidth}{!}{
    \begin{tabular}{c|c|ccc|c}
        \toprule[1pt]
        \textbf{Method} & \textbf{Protocol} & \textbf{Fog} & \textbf{Rain} & \textbf{Snow} & \textbf{Avg.} \\
        \toprule[0.5pt]
        Source & Frozen & 67.72 & 75.99 & 59.32 & 67.68 \\
        \toprule[0.5pt]
        MOS & Independent & 69.11 & 77.79 & 62.72 & 69.87 \\
        MOS & Continual & 69.11 & 76.53 & 60.85 & 68.83 \\
        \rowcolor[HTML]{E6F1FF} Ours (\textbf{FIND}) & Independent & 71.27 & 78.23 & 64.89 & 71.46 \\
        \rowcolor[HTML]{E6F1FF} Ours (\textbf{FIND}) & Continual & 71.27 & 78.04 & 63.96 & 71.09 \\
        \toprule[1pt]
    \end{tabular}
    }
\end{table}

\section{Additional Ablation Study}
\label{appendix:ablation}
In Sec.~\ref{sec:ablation} of the main paper, we analyze the contribution of FIND's components and its sensitivity to selected hyperparameters. In this section, we further examine how changes to the adaptation design and the reliability of its inputs affect detection performance. These studies clarify the conditions under which FIND maintains effective adaptation.

\subsection{Hyperparameter Robustness}
We further evaluate the sensitivity of our method to variations in key hyperparameters: (1) the loss weights $\beta_{\text{st}}$ and $\beta_{\text{comp}}$, (2) the stability mask decay factor $\tau$ in Equ.~(\ref{equ:init}) and (\ref{equ:stability}), (3) the threshold initialization $\eta_{th}^{(0)}$ for Invariance Stream selection.

\begin{figure}[!t]
    \centering
    \includegraphics[width=\linewidth]{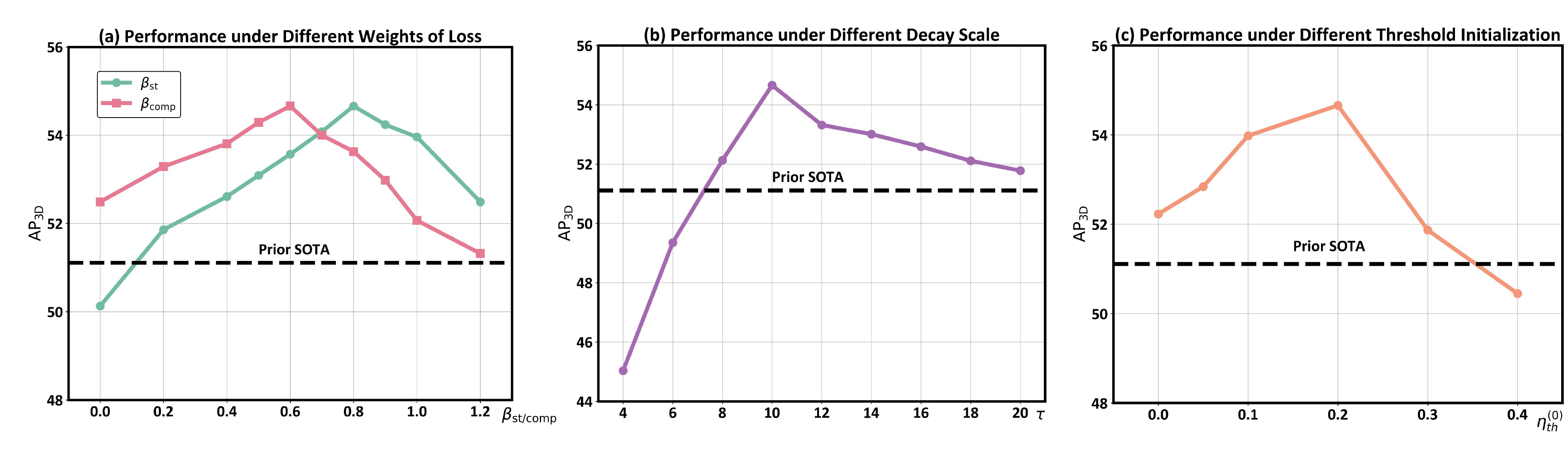}
    \caption{(a) Performance with varying weights $\beta_{\text{st}}$ of $\mathcal{L}_{\text{st}}$ and $\beta_{\text{comp}}$ of $\mathcal{L}_{\text{comp}}$. (b) Performance with different decay scale factor $\tau$ for stability mask. (c) Performance with distinct threshold initializations of Invariant Stream.}
    \label{fig:hyper}
\end{figure}

As shown in Fig.~\ref{fig:hyper}(a), our method yields consistent performance gains across a broad range of $\beta_{\text{st}}$ and $\beta_{\text{comp}}$ values. However, increasing $\beta_{\text{st}}$ beyond the optimal range may degrade the filtering module by inducing a collapse to trivial solutions, whereas an excessive $\beta_{\text{comp}}$ diminishes stability guidance, leading to reduced model performance.

Regarding $\tau$, Fig.~\ref{fig:hyper}(b) demonstrates that excessively low values cause degeneration into a full-band stop filter discarding domain-invariant components, whereas high values render the filtering ineffective. Empirically, we adopt $\tau=10$ as the default setting.

For $\eta_{th}^{(0)}$ as shown in Fig.~\ref{fig:hyper}(c), a low value fails to effectively filter specific components, resulting in suboptimal pseudo-labels from the Invariant Stream. In contrast, a high initialization disrupts the geometric structure by over-utilizing incomplete filters during the initial adaptation stage.

\subsection{Filtering Mechanism Analysis}
In this subsection, we compare B-Spline filters against conventional filtering mechanisms, including rigid band-pass filters and polynomial filters under global control (defined as $h(\lambda)=\sum_{k=0}^K c_k\lambda^k$). Additionally, we analyze the impact of the B-Spline hyperparameters, $N$ and $K$, on the filter response functions.

\textbf{Filter Comparison.} To evaluate rigid filters, we apply low-pass and high-pass filters to the Invariant Stream and Specific Stream, respectively. Formally, we define:
\begin{equation}
    h_{\text{in}}(\lambda)=\mathbb{I}(\lambda\le c),\; h_{\text{sp}}(\lambda)=\mathbb{I}(\lambda>c),
\end{equation}
where $c$ represents the cutoff threshold parameter. Similarly, for global polynomial filters, we adopt the following initialization:
\begin{equation}
    h_{\text{in}}(\lambda)=-(\lambda/2)^K+1,\; h_{\text{sp}}(\lambda)=(\lambda/2)^K.
\end{equation}
The results presented in Table~\ref{tab:filter} demonstrate that rigid filters disrupt the geometric structure of objects, leading to performance degradation. Meanwhile, polynomial filters with global control yield suboptimal results due to their lack of local adaptivity.

\begin{table*}[t]
    \centering
    \caption{Comparisons with different filter mechanisms on nuScenes$\rightarrow$KITTI and Waymo$\rightarrow$KITTI scenarios. We report {$\text{AP}_{\text{BEV}} / \text{AP}_{\text{3D}}$} of Car category in moderate difficulty. "Source" denotes model performance with no filters or adaptation. "Rigid" and "Global" denote model performance with rigid band-pass filters and polynomial filters under global control, respectively. We highlight the best results with \textbf{bold}.}
    \resizebox{0.9\textwidth}{!}{
    \begin{tabular}{c|c|c|c|c}
    \toprule[1pt]
        \multirow{2}{*}{\textbf{Filter}} &
        \multicolumn{2}{c|}{\textbf{nuScenes $\rightarrow$ KITTI}} &
        \multicolumn{2}{c}{\textbf{Waymo $\rightarrow$ KITTI}} \\
        \cmidrule(lr){2-3} \cmidrule(lr){4-5} &
        \textbf{$\textbf{AP}_{\textbf{BEV}} / \textbf{AP}_{\textbf{3D}}$} &
        \textbf{Closed Gap} &
        \textbf{$\textbf{AP}_{\textbf{BEV}} / \textbf{AP}_{\textbf{3D}}$} &
        \textbf{Closed Gap} \\
    \toprule
        Source  & 45.50 / 13.78 & - & 65.64 / 26.74 & - \\
        Rigid & 54.31 / 32.56 & $+$23.31\% / $+$31.47\%  & 66.31 / 39.10 & $+$3.80\% / $+$26.46\%  \\
        Global & 69.72 / 48.38 & $+$64.09\% / $+$57.99\% & 72.55 / 50.21 & $+$39.15\% / $+$50.25\% \\
        \rowcolor[HTML]{E6F1FF} Ours (\textbf{B-Spline}) & \textbf{75.21 / 54.66} & \textbf{$+$78.62\% / $+$68.51\%} & \textbf{82.35 / 67.09} & \textbf{$+$94.67\% / $+$86.38\%} \\
    \toprule[1pt]
    \end{tabular}
    }
    \label{tab:filter}
\end{table*}

\begin{figure}[!t]
    \centering
    \includegraphics[width=\linewidth]{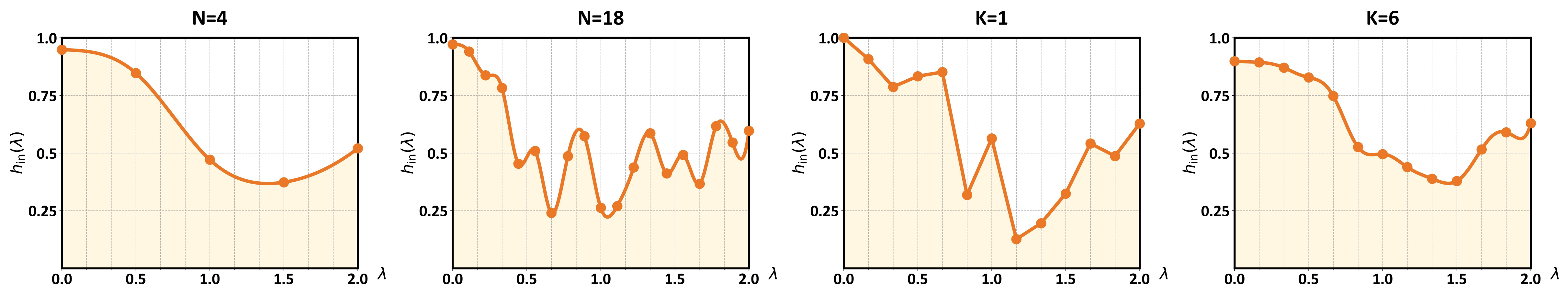}
    \caption{Visualization of B-Spline filter response functions within Invariant Stream under 4 cases:$N=4, N=18,K=1,K=6$. The response magnitudes range from 0 to 1, representing filtering strength, where higher values indicate greater feature preservation.}
    \label{fig:visual}
\end{figure}

\begin{figure}[!t]
    \centering
    \includegraphics[width=0.8\linewidth]{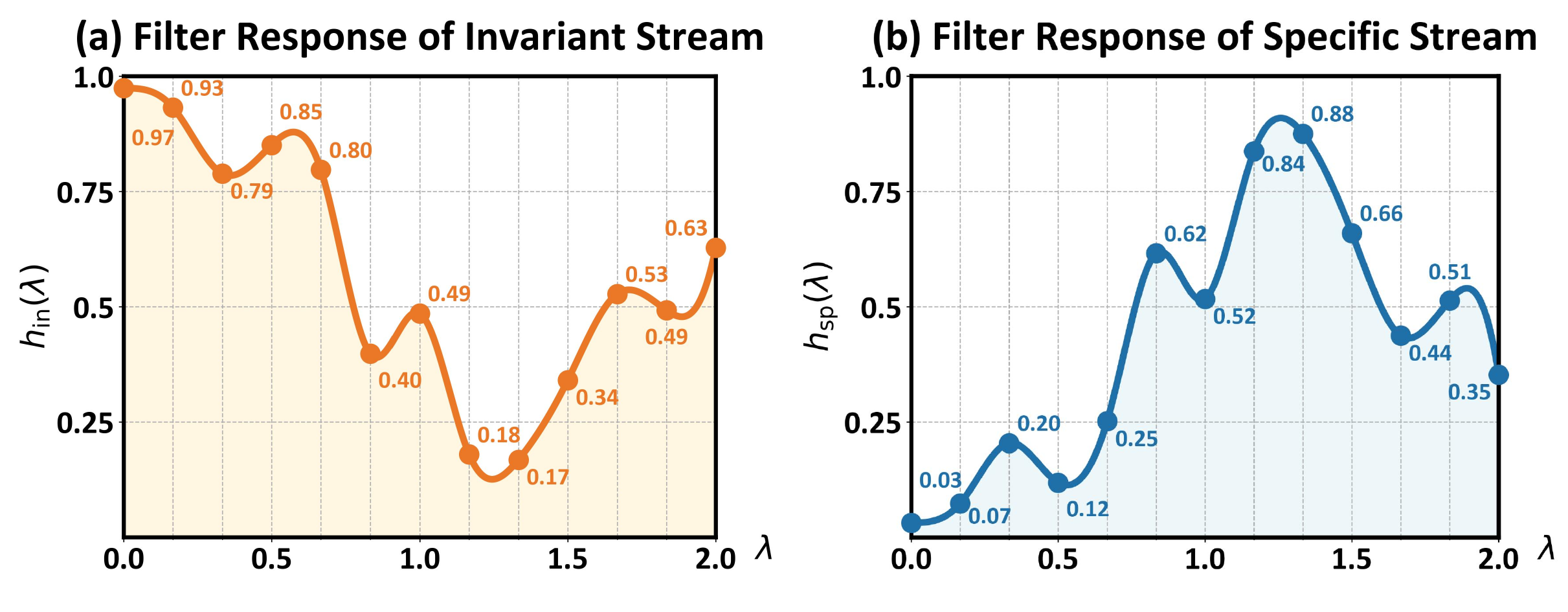}
    \caption{Visualization of adapted B-Spline filter response functions within Invariant Stream and Specific Stream. The response magnitudes range from 0 to 1, representing filtering strength, where higher values indicate greater feature preservation.}
    \label{fig:filter}
\end{figure}

\textbf{Response Function Analysis.} To visually demonstrate the influence of $N$ and $K$ on the filter response function, we conduct image analysis of the filter response function $h_{\text{in}}(\lambda)$ of the Invariant Stream.
As illustrated in Fig.~\ref{fig:visual}, using fewer bins ($N=4$) yields an overly smoothed curve lacking the spectral resolution to isolate narrow-band invariant components. Conversely, an excessive bin count ($N=18$) introduces oscillatory instability, suggesting overfitting to high-frequency domain-specific variations. Regarding the basis order, setting $K=1$ results in a piecewise linear function with derivative discontinuities that compromise the smoothness required for stable filtering. While a higher order ($K=6$) improves smoothness, it expands the local support range, causing spectral leakage that blurs adjacent frequency bands.
Based on these visualizations, we empirically adopt $N=12$ and $K=3$ as the optimal configuration, whose corresponding response filters are shown in Fig.~\ref{fig:filter}, balancing resolution and stability.

\subsection{Robustness to Errors in Predicted Bounding Boxes}
\label{appendix:box-robustness}

Since semantic-aware graph construction uses predicted bounding boxes to select object regions, inaccurate predictions can affect the point clouds supplied to spectral filtering. We evaluate three controlled perturbations while keeping all other settings fixed. Localization noise shifts box centers by up to 5\% of their dimensions, perturbs box dimensions by $\pm5\%$, and changes yaw by $\pm5^{\circ}$. Missed detections remove 10\% of the boxes, while false detections insert 10\% additional boxes. Tab.~\ref{tab:box-robustness} reports the resulting detection performance on nuScenes$\rightarrow$KITTI and Waymo$\rightarrow$KITTI-C.

The three perturbations lead to limited performance degradation within the tested ranges, with missed boxes causing the largest drop. Missed boxes reduce the number of available object graphs, whereas shifted or false boxes introduce incomplete geometry or background points into the graphs. The remaining correct boxes can still provide adaptation signals, and detector updates can in turn improve the boxes used for subsequent batches, offering a potential self-correcting process. This interpretation is consistent with the observed tolerance to moderate box errors. However, with missed boxes, FIND falls slightly below MOS on Waymo$\rightarrow$KITTI-C, showing that its advantage can diminish when valid object regions become less available.

\begin{table}[t]
    \centering
    \caption{Robustness to errors in predicted boxes used for graph construction. We report $\text{AP}_{\text{3D}}$ of the Car category at moderate detection difficulty; Waymo$\rightarrow$KITTI-C is averaged over eight corruption types at heavy severity. \textbf{Gap} denotes the difference in $\text{AP}_{\text{3D}}$ relative to FIND without added perturbations. We highlight the best results with \textbf{bold}.}
    \label{tab:box-robustness}
    \resizebox{0.8\textwidth}{!}{
    \begin{tabular}{l|cc|cc}
        \toprule[1pt]
        \multirow{2}{*}{\textbf{Method}} & \multicolumn{2}{c|}{\textbf{nuScenes $\rightarrow$ KITTI}} & \multicolumn{2}{c}{\textbf{Waymo $\rightarrow$ KITTI-C}} \\
        \cmidrule(lr){2-3} \cmidrule(lr){4-5}
        & $\text{AP}_{\text{3D}}$ & \textbf{Gap} & $\text{AP}_{\text{3D}}$ & \textbf{Gap} \\
        \toprule[0.5pt]
        MOS & 51.11 & $-$3.55 & 32.62 & $-$1.37 \\
        \toprule[0.5pt]
        \rowcolor[HTML]{E6F1FF} Ours (\textbf{FIND}) & \textbf{54.66} & 0.00 & \textbf{33.99} & 0.00 \\
        \quad + Localization noise & 54.22 & $-$0.44 & 33.13 & $-$0.86 \\
        \quad + Missed boxes & 52.08 & $-$2.58 & 32.39 & $-$1.60 \\
        \quad + False boxes & 54.30 & $-$0.36 & 33.46 & $-$0.53 \\
        \toprule[1pt]
    \end{tabular}
    }
\end{table}

\clearpage

\end{document}